\documentclass[10pt,journal,comsoc]{IEEEtran}
\usepackage{amsbsy,amsmath,amssymb,epsfig,bbm,mathrsfs,multirow,amsthm}
\usepackage[ruled, linesnumbered]{algorithm2e} 
\usepackage{algorithm2e, setspace}
\usepackage{color}
\usepackage{algpseudocode}
\usepackage{graphicx,caption,subcaption,array,multirow}
\usepackage{makecell}
\usepackage[hidelinks]{hyperref}
\usepackage{booktabs} 
\usepackage{cite}

\usepackage{setspace}

\DeclareMathAlphabet{\mathpzc}{OT1}{pzc}{m}{it}
\DeclareMathOperator*{\argmax}{\arg\!\max}

\begin{document}

\title{ Transferable Deep Reinforcement Learning Framework for Autonomous Vehicles with Joint Radar-Data Communications}
\author{Nguyen Quang Hieu, Dinh Thai Hoang, Dusit Niyato, \\ Ping Wang, Dong In Kim, and Chau Yuen
\thanks{N. Q. Hieu and D. T. Hoang are with University of Technology Sydney, NSW, Australia, emails: \{hieu.nguyen-1@student.uts.edu.au, hoang.dinh@uts.edu.au\}.
D. Niyato is with with Nanyang Technological University, Singapore, email: dniyato@ntu.edu.sg.
P. Wang is with York University, Canada, email: pingw@yorku.ca.
D. I. Kim is with Sungkyunkwan University, Korea, email: dikim@skku.ac.kr.
C. Yuen is with Singapore University of Technology and Design, Singapore, email: yuenchau@sutd.edu.sg.
}
\thanks{
This research was supported in part by the Australian Research Council under the DECRA project DE210100651.
This research is supported in part by the programme DesCartes - the National Research Foundation, Prime Minister’s Office, Singapore under its Campus for Research Excellence and Technological Enterprise (CREATE) programme and under its Emerging Areas Research Projects (EARP) Funding Initiative, Alibaba Group through Alibaba Innovative Research (AIR) Program and Alibaba-NTU Singapore Joint Research Institute (JRI), the National Research Foundation, Singapore under the AI Singapore Programme (AISG) (AISG2-RP-2020-019), and Singapore Ministry of Education (MOE) Tier 1 (RG16/20).
This research was supported in part by the National Research Foundation of Korea (NRF) Grant funded by the Korean Government (MSIT) under Grant 2021R1A2C2007638 and the MSIT under the ICT Creative Consilience program (IITP-2020-0-01821) supervised by the IITP.
This research is supported by A*STAR under its RIE2020 Advanced Manufacturing and Engineering (AME) Industry Alignment Fund – Pre Positioning (IAF-PP) (Grant No. A19D6a0053. 
Any opinions, findings and conclusions or recommendations expressed in this material are those of the author(s) and do not reflect the views of A*STAR.

}
}

\maketitle
\begin{abstract}
Autonomous Vehicles (AVs) are required to operate safely and efficiently in dynamic environments. For this, the AVs equipped with Joint Radar-Communications (JRC) functions can enhance the driving safety by utilizing both radar detection and data communication functions. However, optimizing the performance of the AV system with two different functions under uncertainty and dynamic of surrounding environments is very challenging.
In this work, we first propose an intelligent optimization framework based on the Markov Decision Process (MDP) to help the AV make optimal decisions in selecting JRC operation functions under the dynamic and uncertainty of the surrounding environment. We then develop an effective learning algorithm leveraging recent advances of deep reinforcement learning techniques to find the optimal policy for the AV without requiring any prior information about surrounding environment.
Furthermore, to make our proposed framework more scalable, we develop a Transfer Learning (TL) mechanism that enables the AV to leverage valuable experiences for accelerating the training process when it moves to a new environment. 
Extensive simulations show that the proposed transferable deep reinforcement learning framework reduces the obstacle miss detection probability by the AV up to $67\%$ compared to other conventional deep reinforcement learning approaches.
With the deep reinforcement learning and transfer learning approaches, our proposed solution can find its applications in a wide range of autonomous driving scenarios from driver assistance to full automation transportation.
\end{abstract}


\section{Introduction}

%
Recent years have witnessed a fast development of Autonomous Vehicles (AVs) with advanced driving safety technologies. According to the recent Allied Market Research’s outlook~\cite{avmarket2018}, the autonomous vehicle global market is expected to reach \$556.67 billion by 2026 due to significant demands of road safety. To achieve driving safety, AVs are usually equipped with Artificial Intelligent (AI) softwares, light detection \& ranging (LiDAR), and radar detection technologies~\cite{yurtsever2020}. Furthermore, AVs are recommended to communicate with each other, e.g., using Vehicle-to-Vehicle (V2V) communications, or with roadside infrastructures, e.g., using Vehicle-to-Infrastructure (V2I) communications, to facilitate intelligent traffic management, routing and data analysis~\cite{ma2020, gaudio2020}. Aiming for fully-automated future transportation systems, AVs are expected to be equipped with two main components that are (i) advanced driver-assistance system (ADAS) and (ii) data communication system~\cite{ma2020, luong2020}.

Several research works focusing on vision-based ADAS, i.e., camera and LiDAR, have been proposed to address driving safety problems for  AVs~\cite{maqueda2018, hnewa2020}. With the recent development of computer vision and deep learning, AVs can achieve accurate detection under different requirements of autonomous driving applications~\cite{yurtsever2020}. However, most of these aforementioned works focus on enhancing the detection accuracy without considering the impact of environment (e.g., rain, fog, and snow) on the performance and safety of the AVs. Radar, in contrast, can be equipped on AVs and work in such weather conditions with effective detection ability ranging from few meters to more than 200 meters~\cite{ma2020}. 

AVs equipped with data communication systems can exchange information with roadside units (e.g., base stations and edge computing services) to efficiently navigate under different weather conditions or avoid unwanted traffic congestions~\cite{ma2020}. Motivated by the fact that the radar detection and data communication functions can work effectively together under the same frequency band, e.g., mm-wave band at $71-76$ GHz, many research works have been proposed to use these functions in a joint manner, namely Joint Radar-Communications (JRC). Accordingly, this joint design can lead to significant benefits in terms of size, cost, power consumption, robustness, and performance~\cite{luong2020, yurtsever2020}. Therefore, JRC, with its abilities of enhancing driving safety and facilitating intelligent road management, is expected to  play a key role for future AVs.

Although the development of JRC brings major benefits for AVs, they are facing several technical challenges.
In particular, the radar detection and data communication functions work on the same frequency band, and thus optimizing resource sharing between these two functions needs to be addressed~\cite{luong2020}.
In addition, several environment factors can affect operation of the JRC functions and thus yield a dynamic optimization problem which is hard to be solved~\cite{zang2019,hieu2020}. Under different environment conditions, e.g., communication channel, weather, and traffic density, the AVs should be able to control their JRC functions effectively. 
Another problem is that the AVs are highly mobile. When they travel from the current environment to a new environment which has different data distributions, e.g., different traffic density, weather, and road state distributions, they should be able to adapt their JRC functions to suit the new environment. 

To address these aforementioned problems, in this work, we first formulate the dynamic JRC function selection problem of an AV by using a Markov Decision Process (MDP) framework. The MDP framework enables the AV to make decisions by considering the current state of surrounding environment. Then, we propose a deep reinforcement learning algorithm to help the AV learn the environment dynamics and make optimal JRC decisions without requiring any prior information from the surrounding environment.  
Furthermore, we propose a Transfer Learning (TL) approach that enables the AV to  learn quickly an optimal policy when it has to travel to a new environment. This approach also enables an AV to share useful knowledge from a learned environment with other AVs that want to learn optimal policies from this environment. 
The contributions of this paper can be summarized as follows.
\begin{itemize}
\item We propose an MDP framework for the joint radar-communications system in AVs. 
With the MDP framework, the AV can adaptively select the radar detection function or data communication function based on the current status of surrounding environment, e.g., weather condition, communication channel state, traffic condition, and driving speed, to maximize the system performance.
\item We develop a highly-effective algorithm based on Double Deep Q-Network (DDQN) to deal with the dynamic and uncertainty of the environment and find the optimal policy for the AV. The DDQN algorithm with the experience replay memory technique and deep neural network as a function approximator can address the slow-convergence problem of conventional reinforcement learning algorithms. This feature is especially useful when the AV is required to quickly obtain the optimal policy under a large number of uncertain environment factors.
\item To make our proposed framework further scalable, we introduce a TL approach, namely Transfer Learning with Demonstrations (TLwD), to help the AV accelerate the training process when the AV travels to a new environment.  By utilizing a small amount of demonstration data from an outer environment, combining with prioritized experience replay~\cite{schaul2015} and multi-step Temporal Difference (TD) learning~\cite{sutton2018} techniques, the AV can obtain an optimal policy faster than that using the conventional deep reinforcement learning approaches in the new environment. To the best of our knowledge, our paper is the first work that considers TL setting for joint radar-communications in autonomous vehicles.
\item Finally, we perform extensive simulations with the aims of not only demonstrating the efficiency of proposed solutions in comparisons with other conventional methods but also providing insightful analytical results for the implementation of our framework. To that end, we take into account parameters of real automotive radar devices~\cite{dham2017}. In addition, we use statistics from a large dataset, namely nuScenes~\cite{nuscenes}, as our simulation parameters. The simulation results show that our proposed TL approach can reduce the miss detection probability of the AV up to $67\%$ and only require a smaller number of training iterations to converge to the optimal policy, compared to the other conventional deep reinforcement learning approaches.

\end{itemize}

The rest of paper is organized as follows. Related works are reviewed in Section~\ref{sec:related-works}.
Sections~\ref{sec:system-model} and~\ref{sec:problem-formulation} describe the system model and problem formulation, respectively. Section~\ref{sec:drl-approach} presents the Q-learning and Double Deep Q-Network algorithms. Then, Transfer Learning with Demonstrations (TLwD) algorithm is presented in Section~\ref{sec:tl-approach}. After that, the evaluation results are discussed in Section~\ref{sec:simulation-results}. Finally, conclusions are drawn in Section~\ref{sec:conclusions}.

\section{Related Works}
\label{sec:related-works}
Since the AV system equipped with JRC functions can use both radar detection and data communication functions using the same hardware platform, these functions share some joint system resources such as antennas, spectrum, and power.
 Consequently, one major problem is to optimize the resource sharing between the radar detection function and data communication function. 
Recently, some resource sharing approaches have been proposed to solve the problem~\cite{kumari2015, ren2019, dokhanchi2017, dokhanchi2019, kumari2019, cao2020, ren2020}. In~\cite{dokhanchi2017}, the authors proposed a JRC system that enables the vehicle to transmit data while estimating distance, Doppler returns and Angles of Arrival (AoA) of target vehicles. To extract these parameters, the authors introduced two processing methods at the receiver based on Fast Fourier Transform and sub-space followed by diversity combining. Simulation results show the trade-off between the Doppler estimation performance and AoA performance of the two proposed methods.
In~\cite{kumari2015}, the authors proposed an effective framework to support radar detection and data transmission for automotive applications. The key idea of this approach is reserving preamble blocks in the IEEE 802.11ad frame for the radar detection mode to estimate ranges and velocities of targets, and at the same time uses data blocks for the data transmission. Simulation results showed that with SNR values greater than $-2$ dB, the radar echo was detected with very high probability. Similarly, the authors in~\cite{kumari2019, dokhanchi2019} exploited the preamble of a communication frame and used it for the radar detection function. For balancing the resource sharing between radar detection and data communication, the authors in~\cite{kumari2019, dokhanchi2019} proposed a fraction parameter in which the frame contains one fraction for data communication and another one for radar detection. In~\cite{dokhanchi2019}, the authors illustrated performance metrics for different fraction parameters, but the optimal value of this parameter was not discussed. 
In~\cite{kumari2019}, the optimal value of the fraction parameter was provided by using convex hull approximation. 
Although the above works are effective in specific circumstances, they are fixed-schedule schemes that are not appropriate to implement in practice. The reason is that the AVs are highly mobile and when they are running on the road, their surrounding environments are uncertain and dynamic. In addition, the performance of these approaches is limited either by the respective communications protocols or inter-vehicular synchronization.

In~\cite{cao2020, ren2019, ren2020}, the authors proposed time division approaches to optimize the resource sharing between radar detection function and data communication function. In~\cite{ren2019}, a time-slotted JRC network with an uncoordinated setup was considered in which the radar detection probability and data communication throughput were studied. The authors considered two performance metrics, i.e., (i) the maximum achievable detection range of the radar given a tolerable false alarm rate and (ii) the network throughput. Radar update rate and network density were also taken into consideration to evaluate the trade-off of the radar performance and communication performance of the system. However, the fundamental trade-off of radar-communications coexistence with flexible configurations remained unexplored.
The authors in~\cite{cao2020} proposed a time allocation approach for a bistatic automotive radar system aiming to maximize the probability of detection under different conditions of the communication channel. Simulation results showed that with different channel conditions, the optimal portions between radar operation time and data communication operation time are different. In~\cite{ren2020}, the authors proposed a time-sharing scheme in which a JRC node can switch between radar mode and communication mode based on the requirements of radar, i.e., maximum detection range. Simulation results showed that longer communication packet length (i.e., slower radar update rate) can achieve better radar range and higher communication throughput. Although the approaches in~\cite{cao2020, ren2020} can exploit the simplicity of switching-based JRC mechanism~\cite{luong2020}, they are either limited by the number of considered targets~\cite{cao2020} or with fixed settings during operation time relied on the radar maximum detection range~\cite{ren2020}. 

It is worth noting that most aforementioned solutions focus on optimal time-sharing between the radar detection function and data communication function without concerning the dynamic of environment. Optimal time-sharing provides resource utilization, e.g., power and bandwidth, but the optimal solutions are fixed during the runtime, that limits practical applications. Alternatively, these approaches require additional information of environment, e.g., channel model, interference model, or arrival distributions of vehicles in advance, which is intractable to obtain, or even infeasible to collect in real-world applications. The aforementioned approaches also neglect the fact that the AV can possibly move to new environments in which the assumption of data distribution in the old environment fails. For example, the AV can travel from a rural area with a low traffic density, to an urban area with a high traffic density. Moreover, other environment factors such as road condition and weather condition may have different distributions in different environments. Therefore, to achieve robustness in driving safety, AV systems should consider different data distributions in variuos environment settings~\cite{kim2017}. 

To the best of our knowledge, our previous work in~\cite{hieu2020} is the first work considering the joint operation of radar and communication for AVs under high dynamic environments such as unfavorable weather conditions, noisy communication channels. However, our previous work has several limitations. Specially, in~\cite{hieu2020}, we considered that the AV only can choose to either transmit data or perform radar detection function. However, in practice, the AV may have multiple operation modes for these two functions based on specific requirements of autonomous driving applications~\cite{dham2017}. Thus, in this work, we consider the AV with more practical operation modes for the JRC functions which can provide more flexible operations and thus be able to yield a better system performance. Furthermore, unlike our previous work~\cite{hieu2020} which can be applied in a specific environment for a particular AV, in this work we aim to develop a more general framework which can help the AV to quickly obtain an optimal policy when it travels to a new environment.

%
%

\section{System Model}
\label{sec:system-model}

\begin{figure}[t]
\centering
\includegraphics[width=1.0\linewidth]{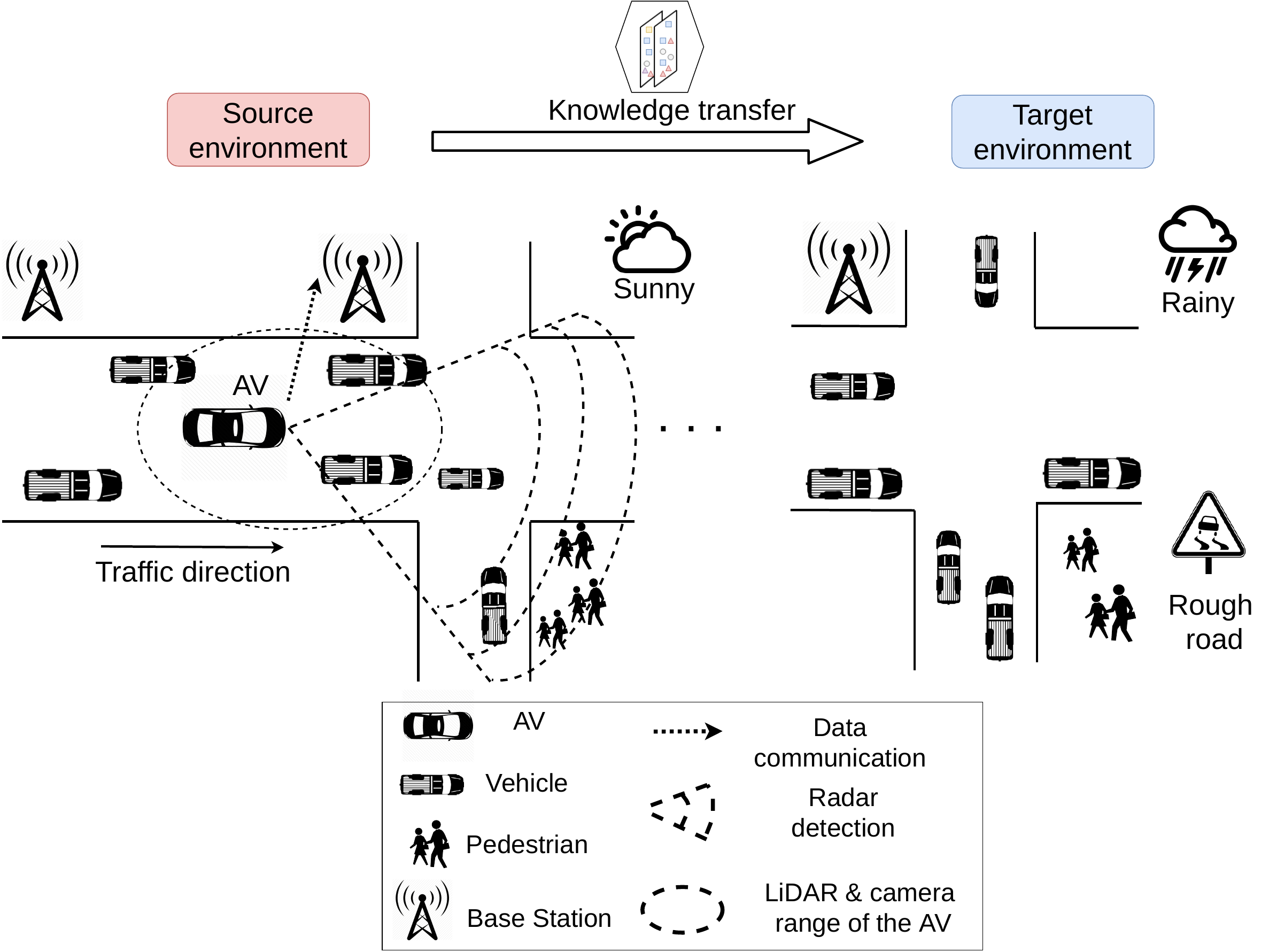}
\caption{An illustration of an AV system equipped with JRC functions.}
\label{fig:system-model}
\end{figure}

We consider an AV system equipped with JRC functions as illustrated in Fig.~\ref{fig:system-model}. In the system model, two environments are considered that are source environment and target environment. The two environments are assumed to be related but different~\cite{zhu2020}. The AV is required to learn efficiently in the source environment and when it travels to the target environment, it should be able to transfer its experiences from the source environment to learn quickly a new policy. Since the two environments are different, the optimal policies learned from these environments are also different~\cite{zhu2020}. Details of the differences between the source environment and target environment are described in the next section.

The AV system in Fig.~\ref{fig:system-model} includes (i) a rate adaptation module for the data communication function and (ii) a Frequency Modulated Continuous Wave (FMCW) radar for the radar detection function. The time horizon is discrete and the AV can choose either radar detection mode or data communication mode at each time step. On the one hand, if the AV uses the data communication mode, it will then select an appropriate data rate $R_l$ to transmit data packets to a nearby Base Station (BS) as illustrated in Fig.~\ref{fig:system-model}. On the other hand, if the AV uses the radar detection mode, it will select a suitable radar mode $B_k$ (e.g., short-range or long-range) to detect objects, e.g., cars and pedestrians in Fig.~\ref{fig:system-model}. Under the dynamics of the environment, e.g., the traffic density, the weather condition, and the road condition, the AV needs to choose a proper radar detection function or data communication function to guarantee safety and transmit data simultaneously.
It is worth noting that the time division-based configuration for the JRC functions benefits from the simplicity and low-cost of implementation~\cite{luong2020}. Both radar detection and data communication functions can be easily implemented using a simple switch without requiring re-design radar and communication waveform~\cite{luong2020}. 
In the following sections, we discuss more details on how these modes are performed at the AV. 

\vspace{-0.3cm}
\subsection{Automotive Joint Radar-Communications Functions}
\label{subsec:jrc-system}
\subsubsection{Data communication function with rate adaption}
The data rate adaptation module enables the AV to adjust its data transmission rates according to the channel quality and thus helps the AV achieve better throughput\cite{yao2016}.
With rate adaptation, the AV can choose one of $L$ data transmission rates from the set $\mathbf{R} = \{R_1, \ldots, R_l, \ldots, R_L \}$, where $R_1 < \cdots < R_l < \cdots < R_L$. These rates can be achieved through different combinations of modulation and coding schemes~\cite{khairnar2014}. The SINR at the BS, denoted by $\eta$, can be then determined by~\cite{yao2016}:

\begin{equation}
\eta = \dfrac{P_R}{\sigma_I + \sigma_N},
\end{equation}
where $P_R$ is the received power at the BS, $\sigma_I$ and $\sigma_N$ are interference power of the other (interfering) signals in the network and random noise power, respectively. For a given SINR, the BS can decode certain rates with target bit error rate (BER). For example, for $l = 1, \ldots, L$, if $\delta_{l-1} \leq \eta < \delta_{l}$ with $\delta_l$ to be the value of SINR, only rates $R_1, R_2, \ldots, R_{l-1}$, can be decoded at the BS~\cite{khairnar2014}. Thus, if the received SINR at the BS is less than $\delta_l$ and the AV chooses a data rate $R_l$ or higher, the transmitted packets will be dropped. An ACK message is used by the BS to notify the AV for the successful transmission.

\subsubsection{Radar detection function with multi-mode FMCW radar}
\begin{figure}
\centering
\includegraphics[width=1.0\linewidth]{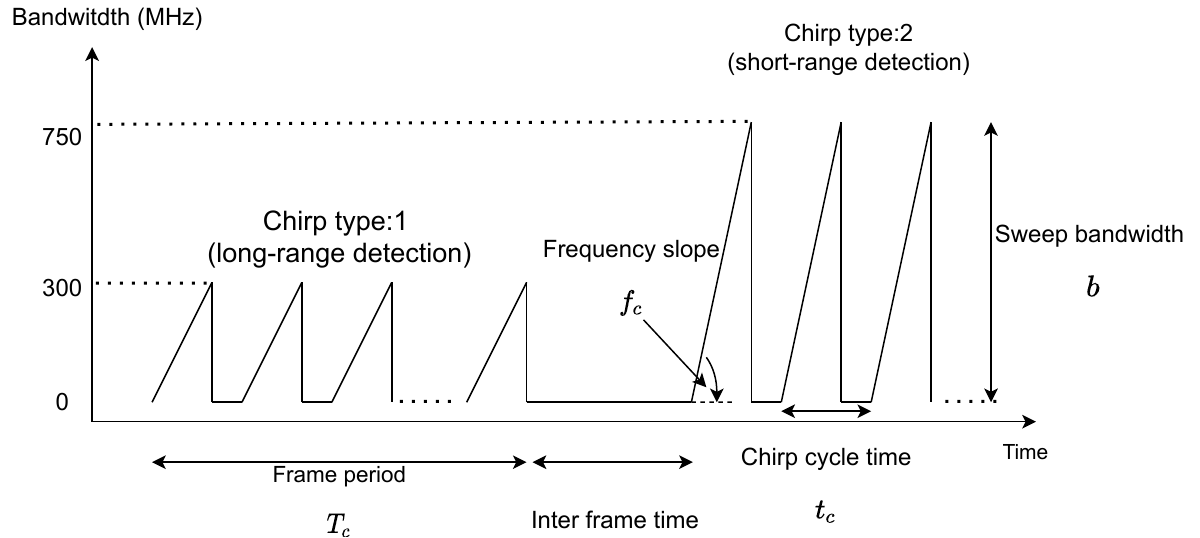}
\caption{Frequency Modulated Continuous Wave (FMCW) mmWave radar with multiple chirp configurations~\cite{dham2017}. 
}
\label{fig:radar-signal}
\end{figure}
FMCW radar can provide multiple detection modes, e.g., short-range detection, mid-range detection, and long-range detection, for the AV to select during its running time~\cite{dham2017}. The multi-mode setting is especially useful when the AV operates in different environments with various requirements. For example, a short-range detection mode can provide a highly accurate range resolution, which can detect nearby objects. Therefore, this mode is very useful in urban environments with high traffic density. 
In particular, each detection mode is equivalent to a specific configuration as illustrated in Fig.~\ref{fig:radar-signal}. For the FMCW radar, radar signals are transmitted as chirps in which the shape of a chirp is decided by sweep bandwidth $b$ and frequency slope $f_c$~\cite{gaudio2020}. This feature enables multi-mode operations on the radar function, e.g., long-range detection with low values of $f_c$ and $b$, and short-range detection with high values of $f_c$ and $b$. 
The relationships between the chirp configuration parameters and the radar performance are given by~\cite{dham2017}:
\begin{equation}
 R_{re} = \dfrac{c_0}{2b}, \text{and }
 R_{max} = \dfrac{c_0 F_{max}}{2 f_c},
\label{eq:range-constraints}
\end{equation}
where $R_{re}$ and $R_{max}$ are the range resolution and maximum detection range of the radar, respectively, $c_0$ is the speed of light, and $F_{max}$ is the maximum intermediate frequency bandwidth supported. 
The range resolution is the minimum distance that the radar can resolve two separated objects.
The maximum detection range is the maximum distance that the AV can detect reflected signals from the far-off objects.
In (\ref{eq:range-constraints}), the values of $c_0$ and $F_{max}$ are constant, while $f_c$ and $b$ can be configured for specific applications, e.g., long/short-range detection~\cite{dham2017}. 

In the system model, the AV can choose one of $K$ radar detection modes from the set $\mathbf{B} = \{B_1, \ldots, B_k, \ldots, B_K\}$, where each mode $B_k$ corresponds to a pair value of sweep bandwidth $b$ and frequency slope $f_c$, i.e., $(b^k, f_c^k)$. As a result, each mode $B_k$ can be illustrated by a pair value of range resolution and maximum detection range, i.e., $(R_{re}^k, R_{max}^k)$~\cite{dham2017}.

\vspace{-0.1cm}
\subsection{Environment Model}
\label{subsec:environment-model}
It is noted that two environments with different dynamics are taken under consideration. However, in this section, we first describe the general characteristics of a specific environment, i.e., the source environment. The details of the differences between the two environments are further explained in Section~\ref{sec:problem-formulation}. 

At each time step, the AV can choose either radar detection mode or data communication mode given the current information of surrounding environment. To model the dynamic of the environment, we consider different factors that have significant influence on the decisions of the AV. These factors are (i) the state of the data communication channel (e.g., good or bad channel condition), (ii) the state of the road (e.g., smooth or rough road), (iii) the state of the weather (e.g., rainy or clear weather), (iv) the speed state of the AV (e.g., high or low speed), and (v) the state of nearby vehicles (e.g., with or without nearby vehicles). The values of these factors can be obtained by the AV's sensing system, e.g., road friction sensor, weather station instrument, speedometer, and cameras.
With the states of these factors, unexpected events can occur. 
We consider an unexpected event to be an event that is out of the range of the AV’s camera and the event can possibly cause collisions with the AV, e.g., a car
coming from another road obscured by a nearby moving vehicle as illustrated
in Fig.~\ref{fig:system-model}.
Also, under a bad weather condition (e.g., rainy or foggy) and a bad road condition (e.g., rough road), if the AV is running in a crowded area (e.g., urban areas), some unexpected events which cannot be detected by the AV's vision-based sensing system (e.g., camera) can occur~\cite{kloeden2001, ustransport}. In this situation, the short-range radar detection mode should be used by the AV to detect more obstructed objects.

To illustrate the dynamic of the environment, we take into account the probability that an unexpected event occurs given current states of the environment. 
Let $c \in \{c_1, \ldots, c_j, \ldots, c_{max}\}$, $r \in \{r_1, \ldots, r_j, \ldots, r_{max}\}$, $w \in \{w_1, \ldots, w_j, \ldots, w_{max}\}$, $v \in \{v_1, \ldots, v_j, \ldots, v_{max}\}$, and $m \in \{m_1, \ldots, m_j,  \ldots, m_{max}\}$ denote the state of the communication channel, the state of the road, the state of the weather, the speed state of the AV, and the state of nearby vehicles, respectively, where the subscript \textit{max} denotes the maximum number of discrete state values. For example, the speed of the AV can be divided into different discrete state values such as $v_1 = 15$ (km/h), $v_2 = 45$ (km/h), and $v_3 = v_{max} = 60$ (km/h). 

Given a particular state of environment, which is a composite of discrete state values  $(r_j, w_j, v_j, m_j)$, let $p_j^f$ denote the probability that an unexpected event occurs at the state value $f_j$ of state $f$ ($f \in \{r, w, v, m\}$). For example, $p_3^v$ expresses the probability that an unexpected event occurs at the current speed state value of the AV is $v_3$, e.g., $v_3 = 60$ (km/h). Clearly, the state value of the communication channel is not considered as a contributing factor to the occurrence of unexpected events.
It is worth noting that these values of $p_j^f$ can be obtained from real-world statistical analysis~\cite{kloeden2001, ustransport}. Then, we can determine the probability that an unexpected event occurs given a particular state $(r_j, w_j, v_j, m_j)$ by using the Bayes' theorem~\cite{deublein2013}. Let $\tau_j^f$ be the probability that state $f$ ($f \in \{r, w, v, m\}$) is at state value $f_j$. The sum of all individual probabilities is equal to 1, i.e.,
\begin{equation}
\underset{j \in \mathcal{J}}{\sum} \tau_j^f = 1,
\end{equation}
where $\mathcal{J} = \{1, 2, \ldots, j_{max}\}$.
By using the Bayes' theorem, the probability of an unexpected event occuring at the state $(r_j, w_j, v_j, m_j)$ is determined by the sum of conditional probabilities as follows~\cite{hieu2020}:

\begin{equation}
p_u = \underset{f \in \mathcal{F}}{\sum} \underset{j \in \mathcal{J}}{\sum} \tau_j^f p_j^f, 
\label{eq:unexpected-ev-prob}
\end{equation}
where $\mathcal{F} = \{r, w, v, m\}$. 

In addition to the factors influencing the probability that an unexpected event occurs as explained above, we need to consider traffic density on the road. The main reason is that in a high-density traffic area, even if the radar system can detect objects, but if the objects are too close to each other, the radar may not be able to differentiate them, and thus unexpected events still can occur. 
Several works have considered traffic density as an important parameter in the field of AVs~\cite{yao2016, nuscenes, robotcar}. In this work, the traffic density is defined as the number of objects in the maximum detection range of the AV.
At time step $t$, given traffic density $\omega$, the miss detection ratio of the AV, denoted by $p_k(t)$, can be calculated as follows:
\begin{equation}
p_k(t) = \dfrac{\omega - N_k(t)}{\omega},
\label{eq:p-miss}
\end{equation}
where $N_k(t)$ is the number of objects that the AV can detect by using the radar detection mode $B_k$ at time step $t$. 
The miss detection ratio represents the ability to detect closely spaced objects as well as far-off objects of the AV. A low miss detection ratio results in a good detection ability of the AV and vice versa~\cite{mazher2020}.
In an area with a high traffic density with closely spaced objects, e.g., urban area, the AV should use the short-range detection mode with a highly accurate range resolution to reduce the miss detection ratio. In contrast, in an area with a low traffic density, the AV should use the long-range detection mode with a low accurate range resolution to detect more objects which are far-off while maintaining the small miss detection ratio.

\vspace{-0.1cm}
\subsection{System Operation and Performance Metrics}
\label{subsec:system-operation}
The operation of the AV system with JRC functions can be described as follows. At each time step, the AV chooses either the radar detection mode or data communication mode. If the AV chooses the data communication mode at time step $t$, it then selects a data rate $R_l$ to transmit $\hat{d}_l(t)$, $\hat{d}_l(t) \in \{\hat{d}_0, \hat{d_1}, \ldots, \hat{d}_{max}\}$, data packets to the nearby BS, where $\hat{d}_{max}$ is the maximum number of data packets that the AV can transmit to the nearby BS. As described in Section~\ref{subsec:jrc-system}, the value of $\hat{d}_l(t)$ depends on the data rate $R_l$ of the AV's transmitter. The AV can transmit the maximum number of data packets $\hat{d}_{max}$ when the channel condition is good and the AV selects the highest data rate $R_L$. Otherwise, the number of packets that can be successfully transmitted to the nearby BS decreases as the channel quality is low.  
If the AV chooses the radar detection mode, it then selects a mode $B_k$ to detect objects on the road. According to the selected mode $B_k$, the miss detection ratio $p_k(t)$ can be obtained as in (\ref{eq:p-miss}).
Here, we assume that the packet error ratio is $0\%$, i.e., no packets loss during transmission. Instead, by introducing the above rate adaptation scheme, our proposed framework still can capture the impacts of communication channel to the number of packets that can be successfully transmitted to the nearby BS, which achieves the same behavior as packet error ratio.

We evaluate the system performance by using two key metrics:  (i) system throughput and (ii) miss detection probability. The system throughput can be determined by the total number of successfully transmitted data packets to the nearby BS, i.e.,
\vspace{-0.2cm}
\begin{equation}
\Psi = \sum_{t=1}^{T}\hat{d}_l(t),
\label{eq:system_throughput}
\end{equation}
where $T$ is the time horizon. The miss detection probability can be calculated as the average miss detection ratio. Thus, the miss detection probability, denoted by $P_k$, can be defined as follows:
\vspace{-0.1cm}
\begin{equation}
P_k = \dfrac{1}{T}\sum_{t=1}^{T} p_k(t).
\label{eq:p_miss_average}
\end{equation}

%
%

Under the dynamic and uncertain environment as well as unexpected events, it is challenging for the AV to optimally select actions to optimize the system performance.
The problem is even more challenging when the action space of the AV, i.e., the number of possible actions in the sets $\mathbf{R}$ and $\mathbf{B}$, increases. 
Thus, all current static optimization tools in~\cite{dokhanchi2019, kumari2019, ren2019, cao2020, ren2020} may not be effective in dealing with these practical issues.
Aforementioned challenges motivate us to develop a dynamic framework which can not only capture the dynamic of environment but also learn incomplete information as well as uncertainty of unexpected events. 
 Accordingly, we formulate the dynamic optimization problem as an MDP and then design a highly-effective deep reinforcement learning algorithm to find the optimal policy without requiring any prior information about the environment.
Furthermore, we propose a new transfer learning approach to leverage the learned experiences of the AV to facilitate the training process in the new environment. 

\vspace{-0.1cm}
\section{Problem Formulation}
\label{sec:problem-formulation}
To deal with the uncertainty of the environment, we use the MDP~\cite{bellman1957} framework to formulate the optimization problem of the system. 
The MDP is defined by a tuple $<\mathcal{S}, \mathcal{A}, \mathcal{T}, \rho>$, where $\mathcal{S}$ is the state space, $\mathcal{A}$ is the action space, $\mathcal{T}$ is the state transition probabilities, and $\rho$ is the immediate reward function of the AV system. Therefore, the source and target environments in our problem are formulated as two MDPs $\mathcal{M}_S  = <\mathcal{S}_S, \mathcal{A}_S, \mathcal{T}_S, \rho_S>$ and $\mathcal{M}_T = <\mathcal{S}_T, \mathcal{A}_T, \mathcal{T}_T, \rho_T>$, respectively. We consider that the two environments share the same state space, action space, and immediate reward function~\cite{zhu2020}. The state transition probabilites of the two environments are different. The MDPs can be rewritten as $\mathcal{M}_S = <\mathcal{S}, \mathcal{A}, \mathcal{T}_S, \rho>$ and $\mathcal{M}_T = <\mathcal{S}, \mathcal{A}, \mathcal{T}_T, \rho>$. 

\vspace{-0.2cm}
\subsection{State Space}
\label{subsec:state-space}

As explained in Section~\ref{subsec:environment-model}, the state space of the AV system is defined as follows:
\begin{multline}
\mathcal{S} = \Big\lbrace (c, r, w, v, m);  c \in \{c_1, c_2, \ldots, c_{max}\},
r \in \{r_1, r_2, \\ \ldots, r_{max}\}, w \in \{w_1, w_2, \ldots, w_{max}\}, 
v \in \{v_1, v_2,\\ \ldots v_{max}\}, \text{and } m \in \{m_1, m_2, \ldots, m_{max}\} \Big\rbrace,
\end{multline}
where $c, r, w, v$, and $m$ represent the state of the communication channel, state of the road, state of the weather, speed state of the AV, and state of nearby vehicles, respectively. 
The state of the system at time step $t$ is defined as $s_t = (c, r, w, v, m) \in \mathcal{S}$. Note that unlike conventional approaches in~\cite{ren2019, cao2020, ren2020} in which additional information of the environment is required, e.g., data distribution of the environment, our approach needs only the instantaneous states of the environmental factors by using AV's sensing system, e.g., road friction sensor, weather station instrument, speedometer, and cameras.

\subsection{Action Space}
\label{subsec:action-space}

As described in Section~\ref{subsec:system-operation}, the possible action at time step $t$ of the AV can be defined as follows: 

\begin{equation}
a_t = \begin{cases}
a_k^r, 1 \leq k \leq K, &\text{if a radar detection mode is selected},  \\
a_l^d, 1 \leq l \leq L, &\text{if a data communication mode is se-} \\ \text{lected}, \\
\end{cases}
\end{equation}
where $a_k^r$ is the action when the AV selects the radar detection mode $B_k$ and $a_l^d$ is the action when the AV selects the data communication mode to transmit data at the rate $R_l$. Thus, the action space of the AV can be defined as $\mathcal{A} = \Big\lbrace a_t; a_t \in \{a_k^r, a_l^d\}\Big\rbrace$. 

\subsection{State Transition Probabilities}
The state transition probabilities of the source and target environments are defined as follows:
\begin{equation}
\mathcal{T}_S = P(s'|p_{u,S}, \omega_S),  \text{and } 
\mathcal{T}_T = P(s'|p_{u,T}, \omega_T),
\end{equation} 
where $p_{u,S}$ and $p_{u,T}$ are the probabilities that an unexpected event occurs as defined in (\ref{eq:unexpected-ev-prob}) of the source environment and target environment, respectively. $\omega_S$ and $\omega_T$ are the traffic density values as defined in (\ref{eq:p-miss}) of the source environment and target environment, respectively. $s'$ is the new state observed by the AV after taking action $a$. Note that the state transition probabilities illustrate the dynamics of environments and they are unknown to the AV. 
  
\subsection{Immediate Reward Function}

At time step $t$ given state $s_t$, the AV selects an action $a_t$ and receives an immediate reward $\rho_t$. The immediate reward $\rho_t$ is designed to maximize jointly the data transmission and radar detection functions of the AV. Thus, the immediate reward can be defined as follows:

\begin{equation}
\rho_t(s_t, a_t) = 
\begin{cases}
\rho_1 \hat{d}_l(t), \text{ if $a_t = a_l^d$, given ${\mathit{\bar{X}}}$}, \\
-\rho_2, \mbox{\ if $a_t = a_l^d$, given $\mathit{X}$}, \\
-\rho_3, \mbox{\ if $a_t = a_k^r$, given $\mathit{\bar{X}}$}, \\
\rho_4(1 - p_k(t)), \mbox{\ if $a_t = a_k^r$, given $\mathit{X}$},
\end{cases}
\label{eq:reward}
\end{equation}
where $(\rho_1, \rho_2, \rho_3, \rho_4)$ are weighting factors, $\mathit{X}$ denotes the occasion when an unexpected event occurs with probability $p_u$ as defined in (\ref{eq:unexpected-ev-prob}) and $\mathit{\bar{X}}$ denotes the occasion when no unexpected event happens. $\hat{d}_l(t)$ is the number of data packets successfully transmitted to a nearby BS, $p_k(t)$ is the miss detection ratio as defined in (\ref{eq:p-miss}). It is worth noting that,  weighting factors $(\rho_1, \rho_2, \rho_3, \rho_4)$ can be adjusted to the requirements of different scenarios. For example, if the AV requires a highly accurate radar detection and the data communication function is secondary, $\rho_3$ and $\rho_4$ can be set at large values while $\rho_1$ and $\rho_2$ can be set at small values.

The immediate reward function of the AV in (\ref{eq:reward}) can be described as follows. 
When there is no unexpected event and the AV selects a data communication mode with data rate $R_l$, the AV receives a positive reward that is proportional to the number of data packets successfully transmitted to the BS (the first condition in (\ref{eq:reward})). When there is an unexpected event and the AV selects a data communication mode, it receives a large negative reward, denoted as $-\rho_2$ (the second condition in (\ref{eq:reward})). When there is no unexpected event, if the AV selects a radar detection mode, it receives a negative reward, denoted as $-\rho_3$ (the third condition in (\ref{eq:reward})). When there is an unexpected event, if the AV selects a radar detection mode, it receives a reward that is inversely proportional to the miss detection ratio $p_k(t)$ (the last condition in (\ref{eq:reward})). 
Note that if the AV does not use the radar detection function for early detection at the beginning of current time slot, it may still obtain the occurrence of the unexpected event at the end of the time slot by using its camera and LiDAR sensors. 

\subsection{Optimization Formulation}
We first formulate the optimization problem of the AV in the source environment. The details of how the experiences of the AV can be transferred to the target environment are described in Section~\ref{sec:tl-approach}.
We formulate a dynamic optimization problem to obtain an optimal policy, denoted by $\pi^*$. Given the current system state, i.e., state of the channel, state of the road, state of the weather, state of the speed, and state of nearby vehicles, the optimal policy determines optimal actions to maximize the average expected return, i.e., the average discounted long-term reward, of the AV system. The optimization problem is then expressed as follows:
\vspace{-0.3cm}
\begin{equation}
\max_{\pi} \ R(\pi) = \lim_{T\to\infty} \dfrac{1}{T} \mathbb{E} \Big\{ \sum_{t=0}^{T} {\gamma^t \rho_{t}(\pi)} \Big\},
\label{eq:discounted-return}
\end{equation}
where $R(\pi)$ is the average expected return under the policy $\pi$, $\rho_{t}(\pi)$ is the immediate reward under policy $\pi$ at time step $t$ as defined in~(\ref{eq:reward}), $T$ is the time horizon, and $\gamma \in (0, 1)$, is the discount factor. The optimal policy $\pi^*$ will allow the AV to make optimal decisions at any state $s_t$, i.e., $a^*_t = \pi^*(s_t)$. 
In practice, it is impossible for the AV to obtain the complete model of the MDP, i.e., state transition probabilities of the MDP are unknown by the AV in advance, and thus model-free reinforcement learning approaches are more suitable for finding the optimal policy for the optimization problem formulated as in (\ref{eq:discounted-return}).
 In the next section, we propose to use Q-learning~\cite{watkins1992} to obtain the optimal policy for the above optimization problem. After that, we propose a method based on deep reinforcement learning, namely Double Deep Q-Network (DDQN)~\cite{hasselt2015}, not only to overcome limitations of Q-learning but also to  learn effectively the optimal policy.

\section{Deep Reinforcement Learning Approach For Automotive Joint Radar- Communication Control}
\label{sec:drl-approach}

\subsection{JRC Optimal Policy With Q-learning}
We first propose to use Q-learning algorithm~\cite{watkins1992} to obtain the optimal policy for the AV system in the source environment. 
In particular, Q-learning algorithm uses a Q-table as a mapping function between states and actions. Each state-action value, i.e., Q-value, in the Q-table is iteratively updated by using Bellman equation as follows:
\begin{equation}
Q(s_t, a_t) = Q(s_t,a_t) + \alpha_t \Big[\rho_t + \gamma \max_{a} Q(s_{t+1}, a)  - Q(s_t, a_t) \Big],
\label{eq:q-learning}
\end{equation}
where $s_t = (c_t, r_t, w_t, v_t)$, $a_t = ( a^r_{k,t}, a^d_{l,t})$, and $\alpha_t$ is the state, action, and learning rate at the time step $t$, respectively. 
When the Q-table is fully updated, the optimal policy for the AV can be obtained by selecting the highest state-action value in the Q-table at each given state, i.e., $\pi^*(s_t) = \argmax_a Q^*(s_t, a)$ where $Q^*(s_t,\cdot)$ is the state values of the Q-table when the Q-table is fully updated. To guarantee the convergence, the learning rate $\alpha_t$ is deterministic, non-negative, and satisfies (\ref{eq:q-learning-convergence})~\cite{watkins1992}:
\begin{equation}
\alpha_t \in (0, 1], \sum_{t=1}^{\infty}\alpha_t = \infty, \mbox{and } \sum_{t=1}^{\infty}(\alpha_t)^2 < \infty.
\label{eq:q-learning-convergence}
\end{equation}

Under the conditions in (\ref{eq:q-learning-convergence}), the Q-learning algorithm converges to the optimal policy with probability one~\cite{watkins1992}.
By using a table as a simple data structure to maintain a mapping function between states and actions, Q-learning usually suffers the curse-of-dimensionality problem~\cite{sutton2018}. For example, when the state and action spaces increase, the size of the Q-table increases quadratically, and thus the time for updating the whole Q-table is longer, meaning that the time for obtaining an optimal policy is also larger.
To overcome the limitations of Q-learning, deep reinforcement learning with nonlinear function approximation, i.e., neural networks, has recently introduced to be a highly-effective solution to overcome limitations of conventional Q-table~\cite{luong2019}. Thus, in the following, we introduce the deep reinforcement learning-based approach to find the optimal policy for the AV.

\subsection{Proposed Deep Reinforcement Learning Approach With Double Deep Q-Network for Optimal JRC Policy}
\begin{figure}
\centering
\includegraphics[width=1.0\linewidth]{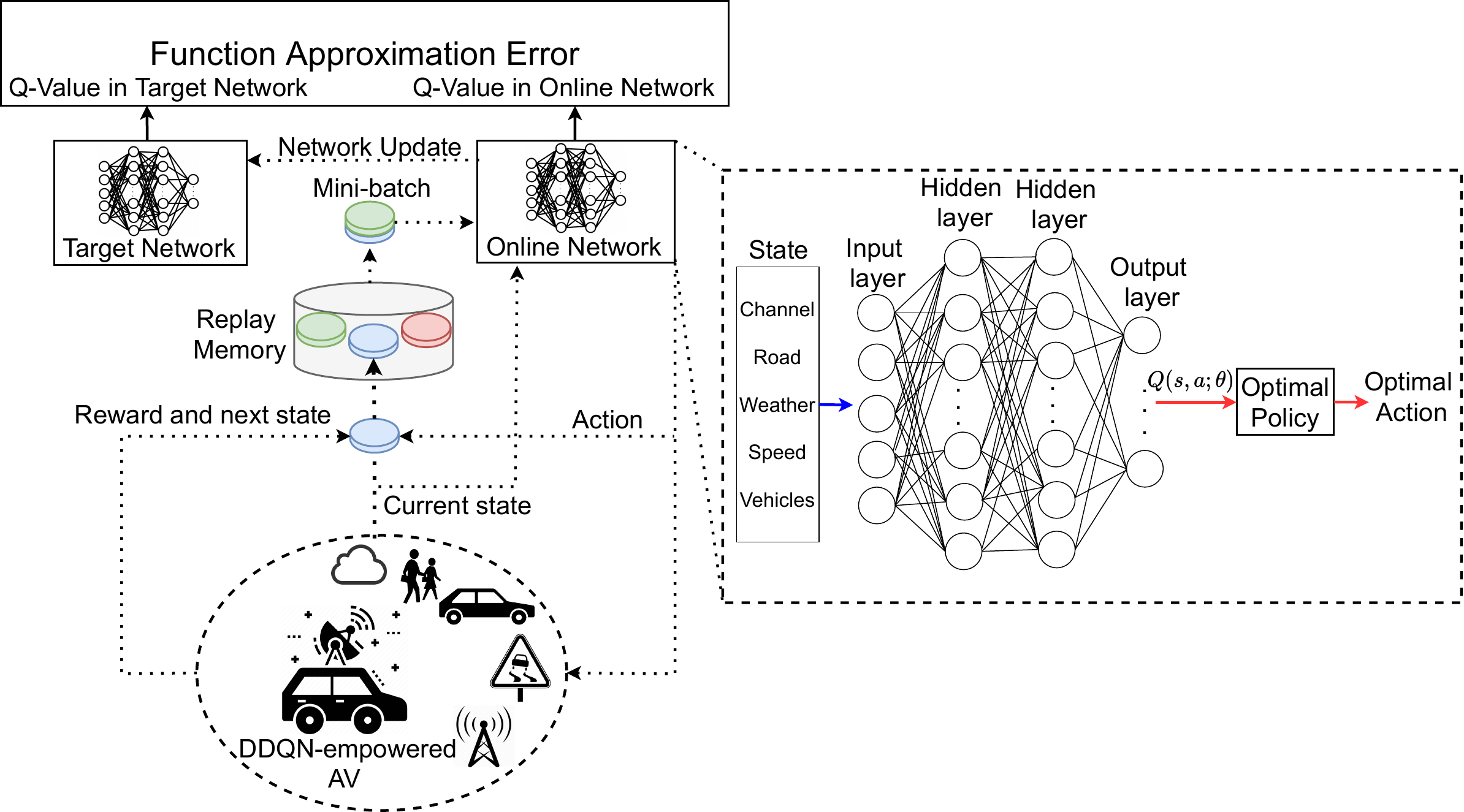}
\caption{Double Deep Q-Network (DDQN) approach for the AV.}
\label{fig:dqn-approach}
\end{figure}

To overcome the limitations of conventional Q-learning~\cite{watkins1992}, Deep Q-Network (DQN)~\cite{mnih2015} has recently introduced. DQN utilizes (i) an online deep neural network (online network) as a nonlinear function approximation, (ii) an experience replay memory as a buffer of historical data, and (iii) a second target deep neural network (target network) as a periodic copy of the online network. The use of the online network as a nonlinear function approximation helps the DQN to overcome the problem related to the large state-action spaces of Q-learning. In addition, an experience replay memory is utilized to store past experiences to speed up the process of updating weights in the online network. Finally, the target network is periodically replaced by the online network and used for stabilizing the training process. 
In this work, we develop a mechanism based on a variation of DQN that is Double DQN (DDQN)~\cite{hasselt2015} not only to find the optimal policy for the AV but also to improve the stability of the training process. In particular, the DDQN and DQN algorithms share the same neural networks' architecture. However, the difference between the DDQN and DQN algorithms is that the calculation of target Q-values in the DDQN algorithm can avoid bias in the action selection~\cite{hasselt2015}. An illustration of our proposed approach is shown in Fig.~\ref{fig:dqn-approach}. In our proposed approach, the DDQN-empowered AV interacts with the environment through its simulator in the training process. When the training process completes in the simulator, the AV can operate in real-world settings with its trained DDQN model. 

\begin{algorithm}[h]
\caption{Double Deep Q-Network-based Algorithm For Finding Optimal Policy}
 \textbf{Inputs}:  \\
 $\mathcal{D}^{re}$: initialize replay memory with capacity $|\mathcal{D}^{re}|$, \\ 
 $\theta$: weights for the initial online network are randomly initialized,  \\
 $\theta'$: weights for the target network are randomly initialized,\\
 $\upsilon$: frequency to update target network, \\
 \For{\mbox{episode} = 1, E}{
 Initialize a new state $s_t \in \mathcal{S}$ \\
 \For{t = 1, T}{
	With probability $\epsilon$, select a random action $a_t$, 
	otherwise select $a_t = \max_a Q(s_t, a; \theta_t)$ \\
	Execute action $a_t$ and observe reward $\rho_t$ and next state $s_{t+1}$ \\
	Replace $s_t \leftarrow s_{t+1}$ \\
	Store transition $(s_t, a_t, \rho_t, s_{t+1})$ in $\mathcal{D}^{re}$ \\
	Sample random mini-batch of $N_b$ transitions $(s_j, a_j, \rho_j, s_{j+1})$ from $\mathcal{D}^{re}$ \\
	Set $y_j^{DDQN}$ as in (\ref{eq:ddqn-target}): 
	$y_j^{DDQN} = \rho_j + \gamma Q(s_{j+1}, \argmax_a  Q(s_{j+1}, a; \theta_t), \theta_t')$ \\
	Perform a gradient descent step on $(y_j^{DDQN} - Q(s_j, a_j, \theta))^2$ according to (\ref{eq:gradient-step}), (\ref{eq:sgd-step}), and (\ref{eq:sgd-update}) \\
	\If{$t \mbox{ mod } \upsilon = 0$}{
   $\theta' \leftarrow \theta$
   } 
   }
 }
 \textbf{Outputs}: $\pi^*(s) = \argmax_a Q^*(s, a; \theta)$. \\
\label{algo:ddqn}
\end{algorithm}

The detailed DDQN algorithm is shown in Algorithm 1 and can be described as follows. The inputs of the network include channel state, road state, weather state, speed state, and surrounding vehicle state. The output is the learned Q-values. 
The experience replay memory, denoted as $\mathcal{D}^{re}$, with capacity $|\mathcal{D}^{re}|$, the weights of the online network and target network, denoted as $\theta$ and $\theta'$, are randomly initialized, i.e., lines 2-5 in Algorithm 1. At each training iteration, the AV selects an action $a_t$ following an $\epsilon$-greedy policy, i.e., selecting $a_t$ randomly with probability $\epsilon$, and otherwise selects $a_t = \max_a Q(s_t, a; \theta)$ (line 9). The AV executes the action $a_t$ and obtains a reward $\rho_t$ and the next state $s_{t+1}$ (line 10). The tuple of $(s_t, a_t, \rho_t, s_{t+1})$ is stored in $\mathcal{D}^{re}$. Next, a mini-batch of transitions $(s_j, a_j, \rho_j, s_{j+1})$ is sampled from $\mathcal{D}^{re}$ to calculate the target Q-values (line 13). 
The target $y_j^{DDQN}$ at time step $t$ is calculated as follows:
\begin{equation}
y_j^{DDQN} = \rho_{j} + \gamma Q(s_{j+1}, \argmax_{a'}  Q(s_{j+1}, a'; \theta_t), \theta_t').
\label{eq:ddqn-target}
\end{equation}

The DDQN algorithm then updates the online network by minimizing the following loss function at time step $t$:
\begin{equation}
\begin{split}
  L_t(\theta_t) = \mathbb{E}_{(s_j, a_j)\sim \mathcal{D}^{re}}\left[(y_j^{DDQN} -  Q(s_j, a_j; \theta_t))^2  \right].
\end{split}
\label{eq:loss_ddqn}
\end{equation}

Differentiating the loss function in (\ref{eq:loss_ddqn}) with respect to the weights of the online network, we have the following gradient:
\vspace{-0.3cm}
\begin{multline}
\nabla_{\theta_t} L_t(\theta_t) = \mathbb{E}_{(s_j, a_j, s_{j+1})\sim \mathcal{D}^{re}} \Bigl[\Bigl(y_j^{DDQN} 
\\  - Q(s_j,a_j;\theta_t)\Bigr)\nabla_{\theta_t} Q(s_j, a_j;\theta_t)\Bigr].
  \label{eq:gradient-step}
\end{multline}

In practice, the gradient in (\ref{eq:gradient-step}) can be efficiently calculated by using Stochastic Gradient Descent (SGD) algorithm~\cite{goodfellow2016}. In general, the loss function in (\ref{eq:loss_ddqn}) is decayed by a sum over training examples of some per-example loss functions. For instance, the negative conditional log-likehood of the training data can be expressed as:
\vspace{-0.1cm}
\begin{equation}
\begin{split}
J_t(\theta_t) &= \quad \mathbb{E}_{(s_j, a_j, s_{j+1}) \sim \mathcal{D}^{re}} L_t(\theta_t) \\ &= \quad \dfrac{1}{|\mathcal{D}^{re}|} \sum_{j=1}^{|\mathcal{D}{re}|} L_t((s_j, a_j, s_{j+1}),\theta_t).
\label{eq:log-likehood}
\end{split}
\end{equation}

 For the additive cost function in (\ref{eq:log-likehood}), the required gradient descent can be computed in the following way:
\vspace{-0.1cm}
\begin{equation}
\nabla_{\theta_t} J_t(\theta_t) = \dfrac{1}{|\mathcal{D}^{re}|} \sum_{j=1}^{|\mathcal{D}^{re}|} \nabla_{\theta} L_t((s_j, a_j, s_{j+1}), \theta_t).
\label{eq:additive-cost-function}
\end{equation}

The computation cost for the operation in (\ref{eq:additive-cost-function}) is $(O|\mathcal{D}^{re}|)$. 
Thus, as the size $|\mathcal{D}^{re}|$ of the replay memory increases, the time to take a single gradient step becomes prohibitively long. 
By using SGD, we can uniformly sample a mini-batch of experiences from the replay memory $\mathcal{D}^{re}$ at each step of the algorithm. In general, the mini-batch size can be set to be relatively small number of experiences, e.g., from one to a few hundreds. As such, the training time is significantly reduced. The estimation of the gradient using SGD is then calculated as follows:
\begin{equation}
g = \dfrac{1}{N_b} \nabla_{\theta_t} \sum_{j=1}^{N_b} L_t(\theta_t),
\label{eq:sgd-step}
\end{equation}
where $N_b$ is the mini-batch size. The SGD algorithm then updates the online network's weights as follows:
\begin{equation}
\theta_{t+1} \leftarrow \theta_t - \nu g,
\label{eq:sgd-update}
\end{equation}
where $\nu$ is the learning rate of the SGD algorithm.
After every $\upsilon$ steps, the target network's weights $\theta'$ are replaced by the online network's weights $\theta$ (lines 16-18 in Algorithm 1). The target network weights remain unchanged between individual updates. 

\subsection{Complexity Analysis for the Double Deep Q-Network Algorithm}
In this work, the DDQN algorithm uses two deep neural networks that are online network and target network. However, we only consider the computations of the online network because the target network is a periodically replaced by the online network and thus it does not require significantly additional computations.
The online network consists of an input layer $L_0$, two fully-connected layers $L_1$ and $L_2$, and an output layer $L_3$. Let $|L_i|$ denote the size, i.e., the number of neurons, of layer $L_i$. The complexity of the online network can be formulated as $|L_0||L_1| + |L_1||L_2| + |L_2||L_3|$. At each training step, a number of training samples, i.e., transitions, are randomly taken from the replay memory and fed to the online network for training. Thus, the total complexity of the training process is $O\Big(TN_b(|L_0||L_1| + |L_1||L_2| + |L_2||L_3|)\Big)$, where $N_b$ is the size of the training batch and $T$ is the total number of training iterations. In our paper, the size of $L_0$ is the number of state features, therefore $|L_0| = 5$. The hidden layers $L_1$ and $L_2$ have 24 neurons each. The output layer consists of $|\mathcal{A}|$ neurons in which each neuron corresponds to an action of the AV as explained in Section~\ref{subsec:action-space}. Clearly, the architecture of the online network is simple. Thus, it can be deployed at the AV system which is usually equipped with sufficient computing resources. In the simulation, we show that our proposed algorithm can converge to the optimal policy much faster than the conventional Q-learning algorithm.

\section{Transfer Learning Approach For Accelerating Learning Process}
\label{sec:tl-approach}

In the previous section, we show that the optimal policy of the source environment can be obtained by using the proposed DDQN algorithm. The optimal policy then can be applied to the target environment with an assumption that the two environments share the same data distribution, i.e., state transition probabilities~\cite{cong2021, zhu2020}. However, this assumption may fail in practice due to the AV's mobility, i.e., it can often travel from an environment to another one with different state transition probabilities. For example, the AV can travel from a rural area with a low traffic density, to an urban area with a high traffic density. In such a case, factors such as road condition and weather condition may have different distributions in different environments. 
Therefore, to achieve a robust and scalable AV system with JRC functions in practice, we need a method that can quickly learn the new optimal policy when the AV moves from one environment to another one. This requirement motivates us to develop the transfer learning approach that can leverage learned experiences to facilitate the learning process in the new environment.

Our considered transfer learning technique can be expressed as follows.
Given two different but related environments that are source environment, modelled with MDP $\mathcal{M}_S$, and target environment, modeled with MDP $\mathcal{M}_T$, the transfer learning process aims to learn an optimal policy $\pi_T^*$ for the target environment by leveraging exterior information $\mathcal{D}_S$ from $\mathcal{M}_S$ as well as interior information $\mathcal{D}_T$ from $\mathcal{M}_T$:
\vspace{-0.1cm}
\begin{equation}
\pi_T^* = \argmax_{\pi} \mathbb{E}_{(s \sim \mathcal{T}_T, a \sim \pi)} \Big[Q_{\pi}(s,a) \Big] , 
\label{eq:transfer-learning}
\end{equation} 
where $\pi = \phi (\mathcal{D}_S \sim \mathcal{M}_S, \mathcal{D}_T \sim \mathcal{M}_T)$. $\mathcal{T}_T$ is the state transition probability in the target environment, $\pi$ is the function mapping from states to actions in the target environment and is approximated using deep neural networks, denoted as $\phi (\cdot)$, trained on both $\mathcal{D}_S$ and $\mathcal{D}_T$~\cite{zhu2020}.

\begin{figure}[t]
\centering
\includegraphics[width=1.0\linewidth]{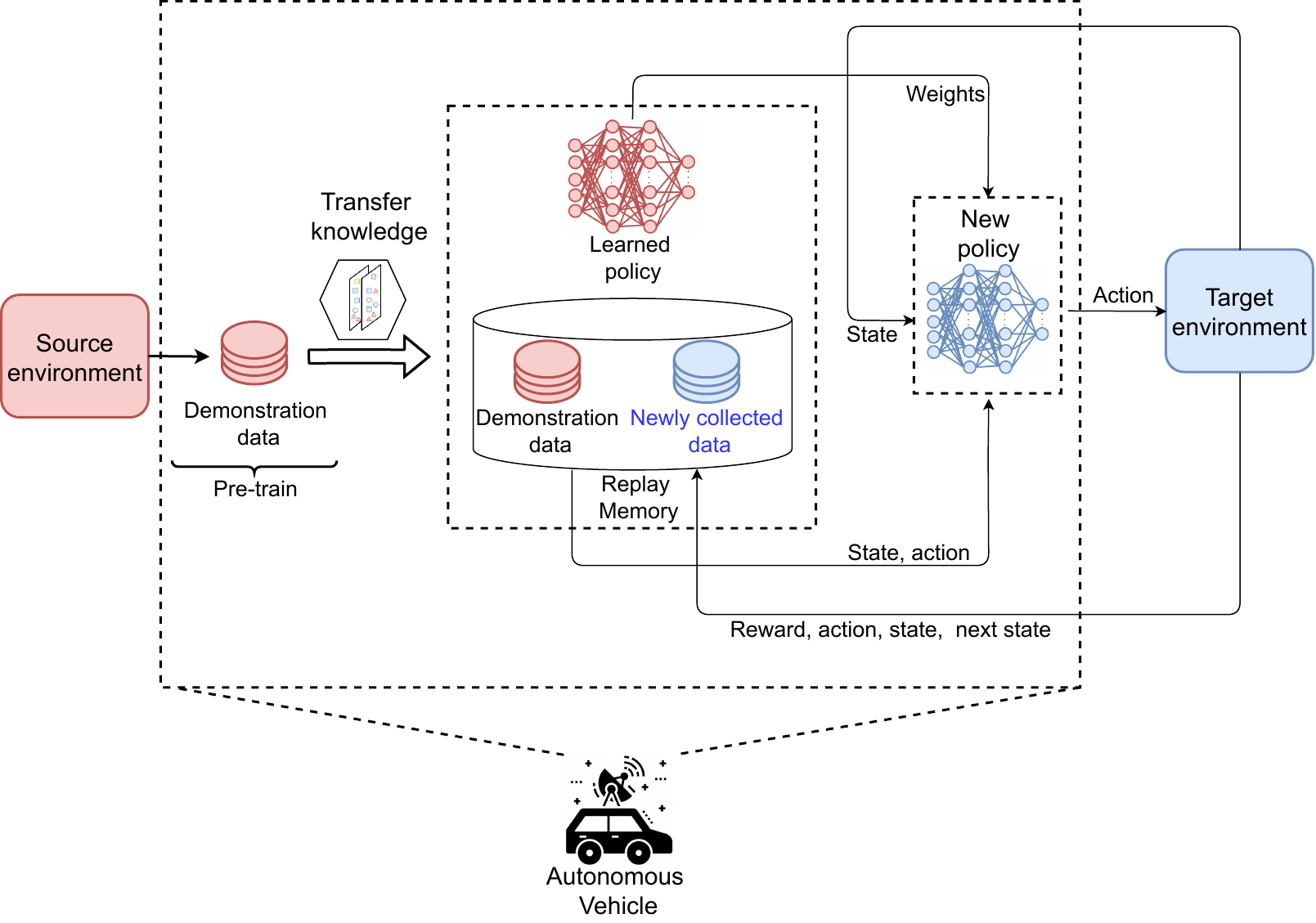}
\caption{Transfer Learning with Demonstrations (TLwD) approach for the AV.}
\label{fig:scenario}
\end{figure}

Several transfer learning approaches have been proposed to solve (\ref{eq:transfer-learning})~\cite{zhu2020}. In autonomous driving, however, collecting exterior information $\mathcal{D}_S$ can be costly and intractable~\cite{weiss2016}. Thus, we need an algorithm that can efficiently accelerate the learning with limited amount of exterior information. In this paper, we develop a transfer learning approach for the AV system, namely Transfer Learning with Demonstrations (TLwD)~\cite{hester2017}. The key idea of TLwD is to initially pre-train on demonstration data $\mathcal{D}_S$ using a combination of temporal difference (TD) and supervised losses. After pre-training, the TLwD algorithm keeps updating its neural networks with a combination of demonstration data $\mathcal{D}_S$ and newly collected data $\mathcal{D}_T$. It is worth noting that both $\mathcal{D}_S$ and $\mathcal{D}_T$ are collections of experiences, i.e., tuples of state, action, reward, and next state, of the AV system in the source and target environments, respectively.
Our proposed TLwD algorithm is illustrated in Fig.~\ref{fig:scenario}.

\begin{algorithm}
\caption{Transfer Learning with Demonstrations}
 \textbf{Input}: \\
 $\mathcal{D}^{re}$: initialized with demonstration dataset, \\ 
 $\theta$: weights for the initial behavior network (random),  \\
 $\theta'$: weights for the target network (random), \\
 $\upsilon$: frequency to update target net, \\
 $T_{pre}$: number of pre-training gradient updates. \\
 \For{$t \in \{1, 2, \ldots, T_{pre} \}$}{
  Sample a mini-batch of $N_b$ transitions from $\mathcal{D}^{re}$ with priotization as in (\ref{eq:prioritized-replay})\\
  Calculate loss $J(Q)$ in (\ref{eq:overall-loss}) using target network \\
  Perform a gradient descent step to update $\theta$ \\
  \If{$t \mbox{ mod } \upsilon = 0$}{
   $\theta' \leftarrow \theta$
   }
 }
 \For{$t \in \{1, 2, \ldots, T \}$}{
  Sample action from behavior policy $a_t \sim \pi^{\epsilon Q_{\theta}}$ \\
  Execute $a_t$ and observe $(s_{t+1}, \rho_t)$ \\
  Store $(s_t, a_t, s_{t+1}, \rho_t)$ into $\mathcal{D}^{re}$, overwrite the oldest collected transition if over capacity\\
  Sample a mini-batch of $N_b$ transitions from $\mathcal{D}^{re}$ with priotization as in (\ref{eq:prioritized-replay})\\
  Calculate loss $J(Q)$ in (\ref{eq:overall-loss}) using target network \\
  Perform a gradient descent step update $\theta$ \\
  \If{$t \mbox{ mod } \upsilon = 0$}{
   $\theta' \leftarrow \theta$
   }
   $s_t \leftarrow s_{t+1}$
 }
 \textbf{Outputs}: $\pi^*_T(s) = \argmax_a Q^*(s, a; \theta)$
\label{algo:tl}
\end{algorithm}

Details of our proposed TLwD approach are illustrated in Algorithm 2 and can be described as follows. 
First, the replay memory $\mathcal{D}^{re}$ is initialized with demonstration data $\mathcal{D}_S$, the weights of online network and target network are randomly initialized (lines 2-6 in Algorithm 2). The inputs of the online network include channel state, road state, weather state, speed state, and state of nearby vehicles. The output is learned Q-values.
The weights of the online network and target networks are pre-trained as follows (lines 7-14). At each pre-training step, a mini-batch of transitions is sampled from $\mathcal{D}^{re}$ with prioritized-experience replay scheme~\cite{schaul2015}. In particular, the probability of sampling transition $i$ can be calculated as follows:
\vspace{-0.1cm}
\begin{equation}
P(i) = \dfrac{p_i^{\alpha}}{\sum_{e \in |\mathcal{D}^{re}|} p_e^{\alpha}},
\label{eq:prioritized-replay}
\end{equation}
where $p_i > 0$ is the priority of transition $i$ and $|\mathcal{D}^{re}|$ is the size of the replay memory. The exponent $\alpha$ determines how much prioritization is used, with $\alpha = 0$ corresponding to the uniform case, i.e., transition $i$ is sampled randomly. The priority of transition $i$ can be calculated by $p_i = |\zeta_i| + \varepsilon$, where $\zeta_i$ is the Temporal Difference (TD) error of transition $i$, $\varepsilon$ is a small positive constant that prevents the edge-case of transitions not being revisited once their TD error is equal to 0.
With the prioritized-experience replay mechanism, the TLwD algorithm can sample important transitions in the demonstration data $\mathcal{D}_S$ and thus can significantly reduce the size of $\mathcal{D}_S$~\cite{schaul2015, hester2017}.
To account for the change in the distribution, updates to the network are weighted with importance sampling weights as follows: 
\begin{equation}
w_i = \left(\dfrac{1}{|\mathcal{D}^{re}|} \cdot \dfrac{1}{P(i)}\right)^{\beta},
\label{eq:weighted-importance-sampling}
\end{equation}
 where $|\mathcal{D}^{re}|$ is the size of the replay buffer and $\beta$ controls the amount of importance sampling with no importance sampling when $\beta = 0$ and full importance sampling when $\beta = 1$. $\beta$ is annealed linearly from $\beta_0$ to 1. 
By using the mini-batch sampled with prioritization, the online neural network weights are updated by minimizing the overall loss $J(Q)$, which is the combination of the DDQN loss, the margin classification loss, multi-step TD error loss, and L2 regulation loss. The overall loss $J(Q)$ is defined as:
\vspace{-0.1cm}
\begin{equation}
J(Q) = J_{DDQN}(Q) + \lambda_1 J_n(Q) + \lambda_2 J_E(Q) + \lambda_3 J_{L2}(Q),
\label{eq:overall-loss} 
\end{equation}
where the component losses are calculated as follows. 
First, the DDQN loss is calculated as:
\begin{align}
J_{DDQN}(Q) = \left(\rho_t + \gamma Q(s_{t+1}, a_{t}^{max}; \theta') - Q(s, a; \theta)\right)^2,
\end{align}
where $\theta$ and $\theta'$ are the parameters of the online network and target network, respectively, and $a_t^{max} = \argmax_a Q(s_{t+1}, a; \theta)$. The multi-step TD loss $J_n(Q)$ is calculated as:
\vspace{-0.1cm}
\begin{align}
J_{n}(Q) = \left(R_t^{(n)} + \gamma^{n} \max_{a} Q(s_{t+n}, a; \theta') - Q(s_t, a_t; \theta)\right)^2,
\label{eq:n-step-td-loss}
\end{align}
where $R_t^{(n)}$ is n-step return from a given state $s_t$ which can be calculated as:
\begin{align}
R_t^{(n)} = \sum_{k=0}^{n-1} \gamma^{k} \rho_{t+k}.
\end{align}
The margin classification loss between data sampled from $\mathcal{D}_S$ and data sampled from $\mathcal{D}_S$ is defined as: 
\begin{align}
J_E(Q) = \max_{a} [Q(s,a) + \Gamma(a_E, a)] - Q(s, a_E), 
\end{align}
where $a_E$ is the action the expert demonstrator, i.e., DDQN with $\mathcal{D}_S$, takes in state $s$ and  $\Gamma(a_E, a)$ is a margin function that is equal to 0 when $a=a_E$ and equal to 1 otherwise. Finally, the L2 regularization loss $J_{L2}(Q)$ is applied to the weights and biases of the neural network to prevent over-fitting on the relatively small demonstration dataset.
In addition, $\lambda_1$, $\lambda_2$, and $\lambda_3$ are weighting parameters of the multi-step TD loss, classification loss, and regulation loss, respectively. 

When the pre-training process completes, the AV starts training process in the target environment (lines 15-26 in Algorithm 2). In particular, the training processes of the TLwD and DDQN algorithms have three main differences that are (i) the structure of replay memory $\mathcal{D}^{re}$, (ii) the prioritized-experience replay mechanism, and (iii) the combined losses $J(Q)$. 
In the next section, we will show the advantages of the three above techniques over the DDQN algorithm in terms of convergence rate.

It is noted that the computational complexity of the proposed TLwD algorithm is similar to that of DDQN algorithm since the two algorithms share the same neural network's architecture. In particular, the complexity of TLwD can be expressed as $O\big(T_* N_b (|L_0||L_1| + |L_1||L_2| + |L_2||L_3|)\big)$, where $T_* = T_{pre} + T$ is the total number time steps of the pre-training process on demonstration data, i.e., $T_{pre}$, and the online training process on newly collected data from the target environment, i.e., $T$.

\section{Performance Evaluation}
\label{sec:simulation-results}
\subsection{Parameter Settings}

In this paper, we consider a scenario in which an AV travels from the source environment to the target environment as shown in Fig.~\ref{fig:system-model}. We consider our problem to be episodic in which the maximum number of time steps in each episode is 300.
The detailed settings for the joint radar-communications functions and environment can be described as follows.

\subsubsection{Joint Radar-Communication Functions Settings}  
\leavevmode
\newline
\textbf{a) Data communication function}:
At time step $t$, if the AV chooses to use the data communication function, it can transmit data at a low data rate or a high data rate, i.e., $a_t \in \{a_1^d, a_2^d\}$.
As described in Section~\ref{subsec:jrc-system}, when the channel condition is good, the AV can successfully transmit $\hat{d}_l(t) = 4$ or $\hat{d}_l(t) = 2$ packets to a nearby BS by using high data rate transmission or low data rate transmission, respectively. However, when the channel condition is bad, the AV can successfully transmit $\hat{d}_l(t) = 0$ or $\hat{d}_l(t) = 2$ packets to a nearby BS by using high data rate transmission or low data rate transmission, respectively. We assume that the communication channel is in a good condition (e.g., low interference) with probability $\tau^c_0$. Therefore, the channel is in a bad condition (e.g., high interference) with probability $1 - \tau^c_0$.

\textbf{b) Radar detection function}:
For the radar detection function, we consider parameters obtained from a real automotive radar sensor from~\cite{dham2017}.
In particular, at a time step $t$, if the AV uses the radar detection function, it can choose the long-range detection mode or short-range detection mode, i.e., $a_t \in \{a_1^r, a_2^r\}$. 
If the AV chooses the long-range detection mode, i.e., $a_t = a_2^r$, its radar operates with the values of sweep bandwidth and frequency slope $(b^1, f_c^1) = (300\mbox{MHz}, 10\mbox{MHz/$\mu s$})$. According to (\ref{eq:range-constraints}), these values of $(b^1, f_c^1)$ are equivalent to the maximum detection range and range resolution $(R_{max}^1, R_{re}^1) = (225m, 0.5m)$, respectively. In other words, the AV can detect objects that are in a range of $225m$ and can distinguish two objects as long as the distance between two objects is more than $0.5m$.
On the other hand, if the AV chooses the short-range detection mode, its radar operates with $(b^2, f_c^2) = (750\mbox{MHz}, 15\mbox{MHz/$\mu s$})$. These values of $(b^2, f_c^2)$ are equivalent to $(R_{max}^2, R_{re}^2) = (45m, 0.2m)$. 
It can be observed from the values of $(R_{max}, R_{re})$ that long-range detection mode results in a low-quality resolution (i.e., high value of $R_{re}$), meaning that it is more difficult to detect two closely spaced objects. In contrast, short-range detection can detect objects in a short distance, i.e., $45m$, with high-quality resolution, i.e., $0.2m$. A summary of parameters used in simulation is shown in Table~\ref{table:parameter-setting}.

\begin{table}[t] 
\renewcommand{\arraystretch}{0.6} 
\centering
\caption{Parameter setting}
\begin{tabular}{*5c}
\toprule
\multicolumn{1}{c|}{Parameter} &  \multicolumn{2}{c|}{Source Env.} & \multicolumn{2}{c}{Target Env.}\\

\midrule
\multicolumn{1}{c|}{$\gamma$}   & \multicolumn{4}{c}{0.99}\\
\multicolumn{1}{c|}{$\epsilon$-greedy}   & \multicolumn{4}{c}{$(1.0 \rightarrow 0.01)$}\\
\multicolumn{1}{c|}{$N_b$}   & \multicolumn{4}{c}{64}\\
\multicolumn{1}{c|}{$(\lambda_1, \lambda_2, \lambda_3)$} & \multicolumn{4}{c}{$(1.0, 1.0, 10^{-5})$}\\
\multicolumn{1}{c|}{$(|\mathcal{D}_S|, |\mathcal{D}^{re}|)$} & \multicolumn{4}{c}{$(20000, 50000)$} \\
\multicolumn{1}{c|}{$(T_{pre}, \upsilon)$} & \multicolumn{4}{c}{$(10000, 300)$} \\
\multicolumn{1}{c|}{$(\alpha, \beta)$} & \multicolumn{4}{c}{$(0.4, 0.6 \rightarrow 1.0)$}\\
\multicolumn{1}{c|}{$(b^1, f_c^1)$} & \multicolumn{4}{c}{$(300\mbox{MHz}, 10\mbox{MHz/$\mu$s)}$} \\
\multicolumn{1}{c|}{$(b^2, f_c^2)$} & \multicolumn{4}{c}{$(750\mbox{MHz}, 15\mbox{MHz/$\mu$s)}$} \\
\multicolumn{1}{c|}{$(\rho_1, \rho_2, \rho_3, \rho_4)$} & \multicolumn{4}{c}{$(1, 100, 1, 50)$} \\
\multicolumn{1}{c|}{Car's length} & \multicolumn{4}{c}{$\mathcal{N}(4.62, 0.18)$} \\
\multicolumn{1}{c|}{Car's width} & \multicolumn{4}{c}{$\mathcal{N}(1.92, 0.08)$} \\
\multicolumn{1}{c|}{Pedestrian's length} & \multicolumn{4}{c}{$\mathcal{N}(0.73, 0.085)$} \\
\multicolumn{1}{c|}{Pedestrian's width} & \multicolumn{4}{c}{$\mathcal{N}(0.68, 0.055)$} \\

\midrule
\multicolumn{1}{c|}{$p_0^f$ $(f \not\in \{v,w\})$} & \multicolumn{2}{c|}{$0.007$} & \multicolumn{2}{c}{$0.004$} \\
\multicolumn{1}{c|}{$p_1^f$ ($f \not\in \{v,w\})$} & \multicolumn{2}{c|}{$0.07$} & \multicolumn{2}{c}{$0.04$} \\
\multicolumn{1}{c|}{$(p_1^v, p_0^v)$} & \multicolumn{2}{c|}{$(0.1, 0.05)$} & \multicolumn{2}{c}{$(0.4, 0.05)$} \\
\multicolumn{1}{c|}{$(p_1^w, p_0^w)$}  & \multicolumn{2}{c|}{$(0.046, 0.005)$} & \multicolumn{2}{c}{$(0.046, 0.005)$} \\
\multicolumn{1}{c|}{$\tau^f_0$} & \multicolumn{2}{c|}{$0.3$} & \multicolumn{2}{c}{$0.7$} \\
\multicolumn{1}{c|}{$\omega$} & \multicolumn{2}{c|}{$54$} & \multicolumn{2}{c}{$27$} \\

\bottomrule
\end{tabular}
\label{table:parameter-setting}
\end{table}

\subsubsection{Environment Settings}
We set the state of the communication channel, state of the road, state of the weather, speed state, and the state of nearby vehicles as $c \in \{0,1 \}$, $r \in \{0,1 \}$, $w \in \{0,1 \}$, $m \in \{0,1 \}$, and $v \in \{0,1 \}$, respectively.
In particular, $c = 1$, $r = 1$, $w = 1$, $v = 1$, and $m = 1$ represent unfavorable conditions, i.e., bad channel condition, rough road, rainy weather, high speed of
the AV, and with nearby vehicles, respectively. In contrast, $c = 0$, $r = 0$, $w = 0$, $v = 0$, and $m = 0$ represent favorable conditions, i.e., good channel condition, smooth road, good weather, low speed of the AV, and without nearby vehicles, respectively. Note that the generalization of the states beyond 0 and 1 is straightforward.
To model the dynamics of environment, the probabilities that an unexpected event occurs at the given speed state and weather state, denoted as $p_j^v$ and $p_j^w$, respectively, in (\ref{eq:unexpected-ev-prob}) are taken from the real-world data in~\cite{kloeden2001, ustransport}, and other probabilities are assumed to be pre-defined. The probabilities that an unexpected event occurs at low speed and high speed of the AV, denoted as $p_0^v$ and $p_1^v$, respectively, are taken from~\cite{kloeden2001} in which if the AV's speed exceeds 60 km/h, the AV's speed is high and otherwise the AV's speed is low. Specifically, the values of  $p_0^v$ and $p_1^v$ are set to be 0.005 and 0.1, respectively. Rain can be considered a common unfavorable weather state, and as in~\cite{ustransport}, we set  $p_1^w = 0.046$ and $p_0^w = 0.005$.

We consider the traffic density and the sizes of objects on the road and the sidewalk are realistic parameters based on nuScenes dataset~\cite{nuscenes}.
nuScenes is a large public dataset that provides a benchmark from measurements of camera, radar and LiDAR sensors on autonomous vehicles.   
nuScenes dataset contains well-classified 23 object classes recorded from 242km traveled in Boston and Singapore.
From~\cite{nuscenes}, we extract distributions of the traffic density, length, and width of the objects and input these parameters into our simulation. In particular, we consider two most common types of objects in nuScenes that are car and pedestrian. The average values of the number of cars and the number of pedestrians in a time step are 20 and 7, respectively~\cite{nuscenes}. Therefore, we set the average value of traffic density, denoted as $\omega$, at 27 objects within the detection range of $(45 \times 45)$ meters in cases the AV uses the short-range radar detection mode. In the case that the AV uses the long-range radar detection mode, the average value of traffic density is set at $27 \times \frac{225}{45} \times \frac{225}{45} = 675$ objects within the detection range of $(225 \times 225)$ meters. In the rest of the paper, we denote $\omega$ as the average number of objects in the detection range of $(45 \times 45)$ meters for the sake of simplicity, and the value of $\omega$ with the long-range radar detection mode can be calculated as mentioned above.
In addition, the car's length, car's width, pedestrian's length, and  pedestrian's width are fitted as the normal distributions that are $\mathcal{N}(4.62, 0.18)$, $\mathcal{N}(1.92, 0.08)$, $\mathcal{N}(0.73, 0.085)$, and $\mathcal{N}(0.68, 0.055)$, respectively~\cite{nuscenes}. 

At each time step $t$ in the simulation, we generate $\omega$ objects that are randomly spaced in the detection range of $(R_{max} \times R_{max})$ meters of the AV. In the simulation, these parameters are represented as a set of 2D arrays and the AV can only obtain these parameters if it uses the radar detection mode in that time step. In addition, we consider that an unexpected event occurs with probability $p_u$ as defined in (\ref{eq:unexpected-ev-prob}). 

\subsubsection{Neural Networks Settings}
For the DDQN algorithm, we adopt parameters based on the common settings for designing neural networks~\cite{mnih2015, goodfellow2016}. Specifically, two fully-connected hidden layers are implemented together with input layer and output layer. The size of the hidden layer is 24. The mini-batch size is set at 64. The size of the experience replay memory is 50,000, and the target network is updated every 300 iterations. For the TLwD parameters, we use the same parameters as in~\cite{hester2017}. The weighting factors between losses in (\ref{eq:overall-loss}) are set at $(\lambda_1, \lambda_2, \lambda_3) = (1, 1, 10^{-5})$. The number of pre-training steps is 10,000. The size of demonstration data and experience replay memory are 20,000 and 50,000, respectively. The parameters used in the simulation are summarized in Table~\ref{table:parameter-setting}.

\subsubsection{Transfer Learning Setting}
In the transfer learning setting, we first let the AV obtain an optimal policy and demonstration data of the source environment. 
Note that the information of source environment can be pre-computed by other AVs and it can be stored at an edge server.
After that, we let the AV continue learning through updating its neural networks' weights in the target environment. 
The differences between the source and target environments are shown in Table~\ref{table:parameter-setting}.
To obtain the optimal policy of the source environment, we evaluate the reward performance of the DDQN algorithm with two baseline schemes that are Q-learning and Round Robin~\cite{hieu2020}. Round Robin is a rule-based mechanism in which the AV selects actions in an iterative manner so that all the actions in the action space are equally selected. As a result, Round Robin can be utilized as a non-learning scheme which yields the stable and lower bound performance.
In the target environment, we evaluate our proposed transfer learning algorithm by comparing with the two baseline schemes that are DDQN and Direct Policy Reuse. The Direct Policy Reuse (DPR) is a mechanism in which the policy learned from the source environment is directly used in the target environment~\cite{cong2021, fernandez2006}. Note that with the DDQN algorithm, the AV learns the environment from scratch, i.e., the weights of neural networks are initialized randomly. The reason that we use the DDQN algorithm as a baseline scheme is to evaluate the advantages of our proposed transfer learning approach in the transfer learning setting. In the rest of the paper, we use the term \textit{DDQN (without TL)} to denote the results obtain by the DDQN algorithm without using any transfer learning technique, i.e., the AV with DDQN algorithm learns the new environment from scratch.

In the next subsection, we evaluate system performance in terms of average reward, system throughput, and miss detection probability. The average reward of an episode can be calculated by averaging all the immediate rewards. The system throughput and miss detection probability are calculated as in (\ref{eq:system_throughput}) and (\ref{eq:p_miss_average}), respectively.

\subsection{Simulation Results}
\subsubsection{Transfer Learning Experiment}
\begin{figure*}[h]
\centering
	\begin{subfigure}[b]{0.38\textwidth}
	 	\centering
	 	\includegraphics[width=\textwidth]{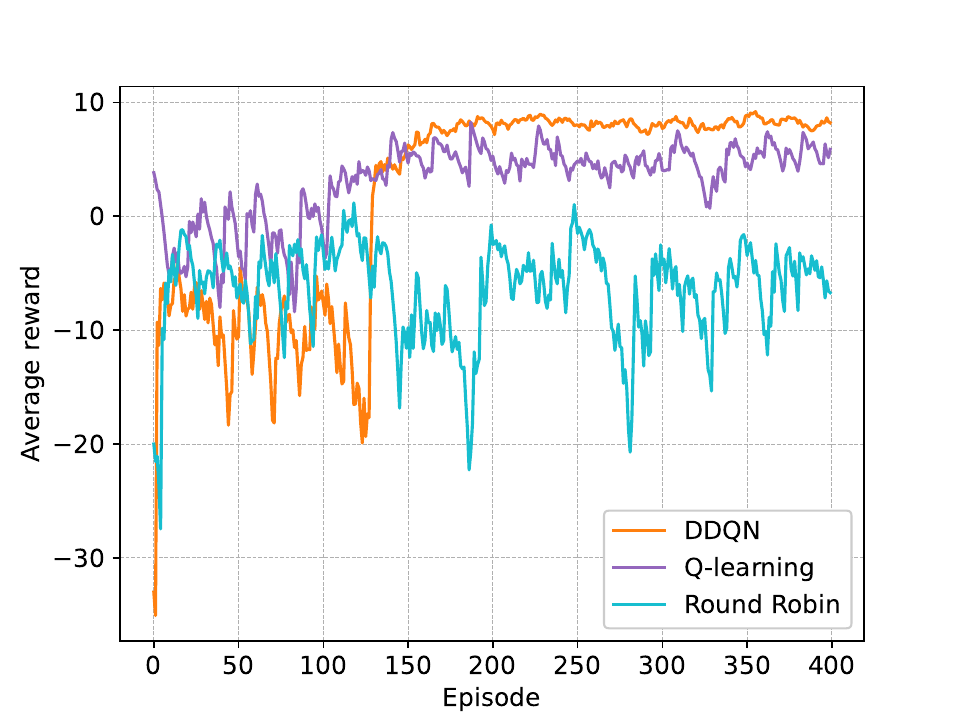}
	 	\caption{}
	 \end{subfigure}
	\begin{subfigure}[b]{0.38\textwidth}
	 	\centering
	 	\includegraphics[width=\textwidth]{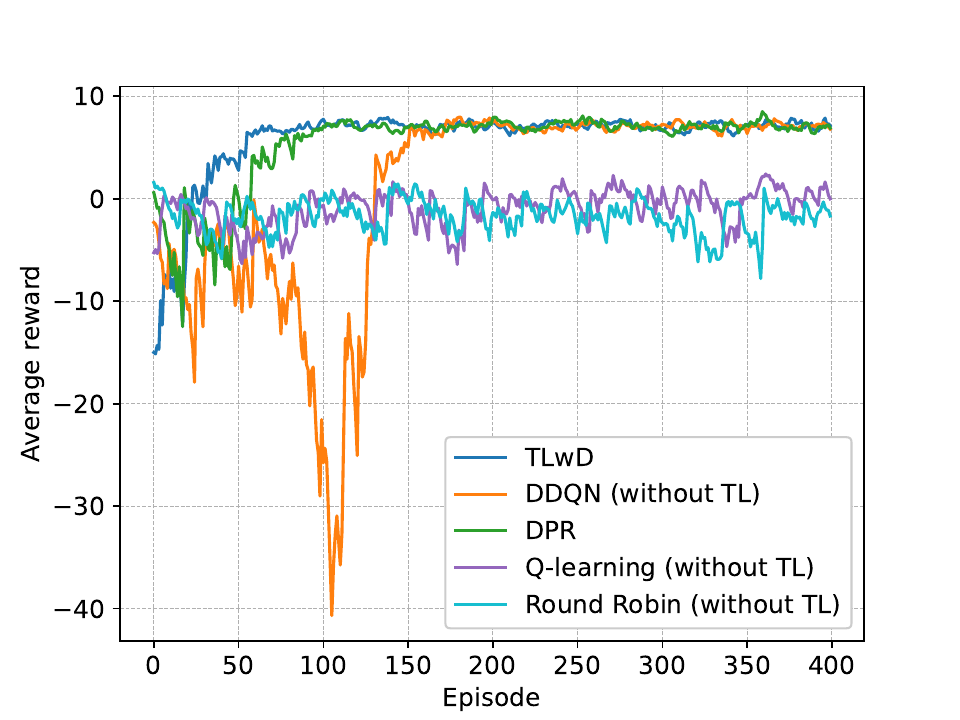}
	 	\caption{}
	 \end{subfigure} 
\caption{(a) Average reward of the DDQN, Q-learning, and Round Robin algorithms in the source environment, and (b) average reward of the TLwD, DPR, DDQN, Q-learning, and Round Robin in the target environment.}
\label{fig:reward-results}
\end{figure*}

First, we let the AV obtain the optimal policy in the source environment by using the DDQN, Q-learning, and Round Robin algorithms. In Fig.~\ref{fig:reward-results}(a), we show the average reward values of the DDQN, Q-learning, and Round Robin algorithms in the source environment.
The result shows that with the same $\epsilon$-greedy scheme, the Q-learning algorithm is unable to obtain the optimal policy while the DDQN algorithm can achieve the optimal policy after approximate 200 episodes.
This means that the Q-learning algorithm has a slow convergence due to the curse-of-dimensionality problem and the considered environment is challenging to learn. With Round Robin algorithm, the actions are selected iteratively and thus the optimal policy cannot be obtained.
When the DDQN algorithm converges to the optimal policy and achieves stable average reward values after 400 episodes, we obtain the weights of the neural network, denoted as $\theta^*_{S}$,  and the demonstration data $\mathcal{D}_S$ from the source environment. The neural network's weights $\theta^*_S$ and demonstration data $\mathcal{D}_S$ are used for initializing DPR and TLwD in the target environment later.
Next, the AV with the obtained optimal policy and demonstration data is used for the learning process in the target environment. In the target environment, all algorithms use the same $\epsilon$-greedy scheme in which the value of $\epsilon$ is reset to 1.0 and is decreased until $\epsilon=0.01$.
In particular, the probability that an unexpected event occurs $p_u$ and the traffic density $\omega$ of the two environments are different (Table~\ref{table:parameter-setting}).
Fig.~\ref{fig:reward-results}(b) shows the average reward values of the proposed TLwD algorithm, compared to DPR, DDQN, Q-learning, and Round Robin algorithms. We observe that the average reward values of TLwD and DPR converge much faster than DDQN. This implies that the transfer learning mechanisms achieve better performance than that of the traditional deep reinforcement learning mechanism. The results also show that the Q-learning and Round Robin algorithms are unable to learn the optimal policy in the target environment. Furthermore, we can observe that although the two learning curves of TLwD and DPR algorithms converge from the 100th episode, the  TLwD algorithm achieves better performance in the first 100 episodes with more stable and higher average rewards. The reason is that the TLwD algorithm leverages the prioritized-experience replay mechanism that helps to sample important transitions in the experience replay memory. Note that the experience replay memory $\mathcal{D}^{re}$ in TLwD algorithm contains two types of data that are the demonstration data obtained from the source environment and the new generated data from the target environment. With prioritized-experience replay mechanism, the AV can sample more important transitions in the demonstration data during early episodes to accelerate the exploration. In the later episodes, the AV can balance the sampling ratio between the demonstration data and new generated data with the sampling probability as defined in (\ref{eq:prioritized-replay}). In addition, the TLwD algorithm also utilizes multi-step TD learning, i.e., n-step TD learning as defined in (\ref{eq:n-step-td-loss}),  that helps to accelerate the training process~\cite{hessel2018}. With the DPR algorithm, the sampling mechanism is the same as that of DDQN algorithm, i.e., the transitions are sampled randomly without using the multi-step learning mechanism. The reason that the DPR outperforms DDQN can be explained as follows. The weights of neural networks in DPR are already trained with data in the source environment and then further updated in the target environment. Although the source and target environments have different state transition probabilities, their data can share some correlation that benefits the weight updating process~\cite{killian2017}.
With DDQN, the weights of neural networks are randomly initialized and trained from scratch.

In the rest of this section, we consider performances of only three algorithms that are TLwD, DPR and DDQN in the target environment since the Q-learning and Round Robin algorithms do not perform well in the transfer learning setting.

\subsubsection{Performance Evaluation}
\begin{figure*}[t]
\centering
	 \begin{subfigure}[b]{0.35\textwidth}
	 	\centering
	 	\includegraphics[width=\textwidth]{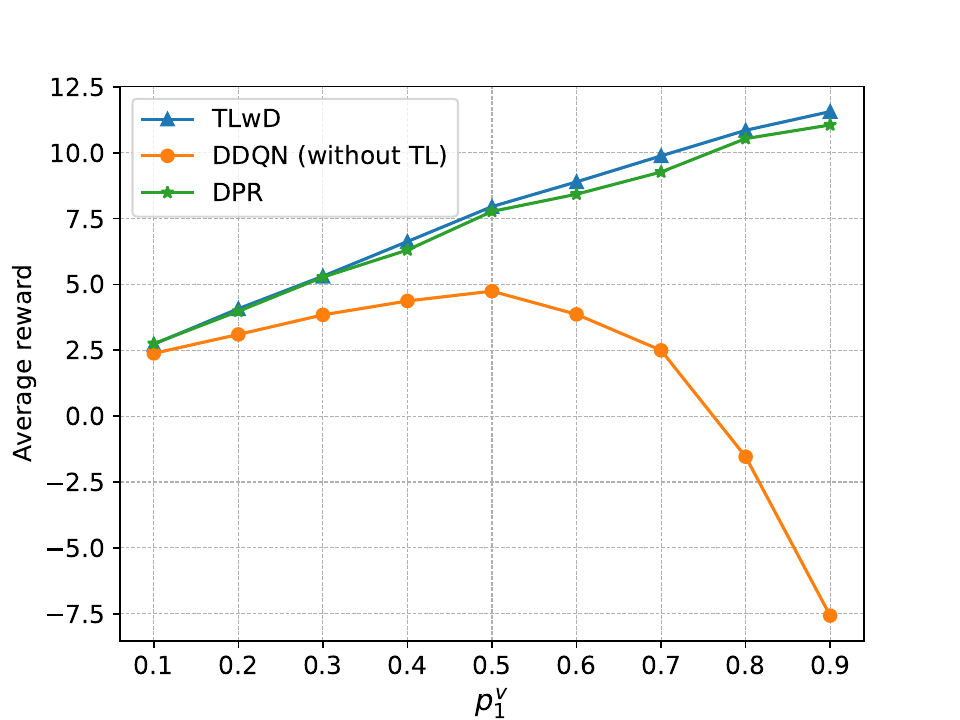}
	 	\caption{}
	 \end{subfigure}
	 \hspace{-0.8cm}
     \begin{subfigure}[b]{0.35\textwidth}
         \centering
         \includegraphics[width=\textwidth]{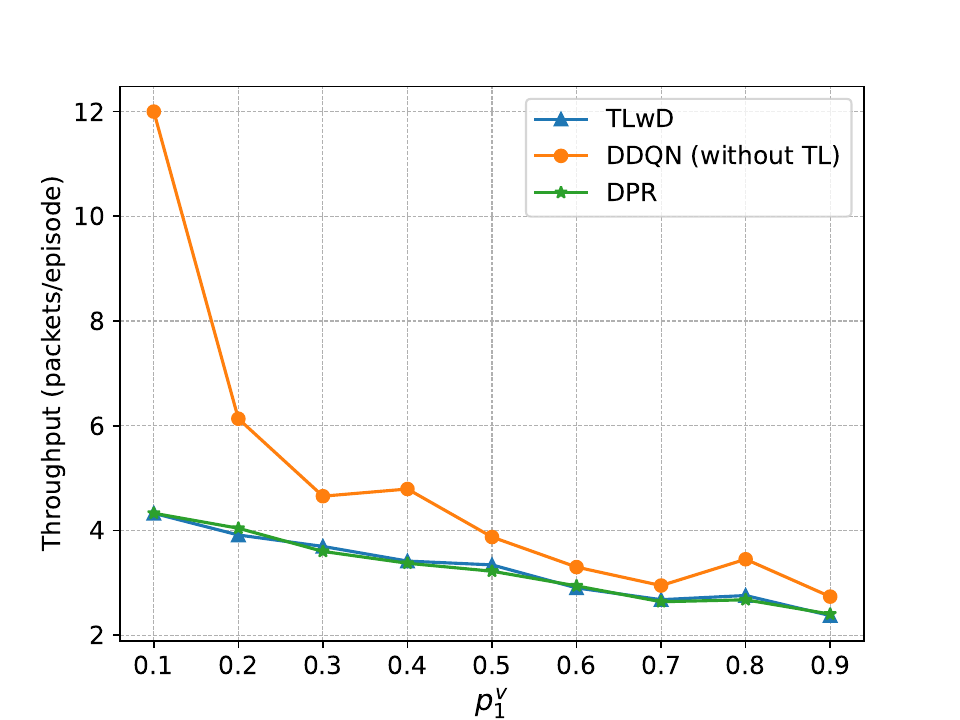}
         \caption{}
     \end{subfigure}
     \hspace{-0.8cm}
     \begin{subfigure}[b]{0.35\textwidth}
         \centering
         \includegraphics[width=\textwidth]{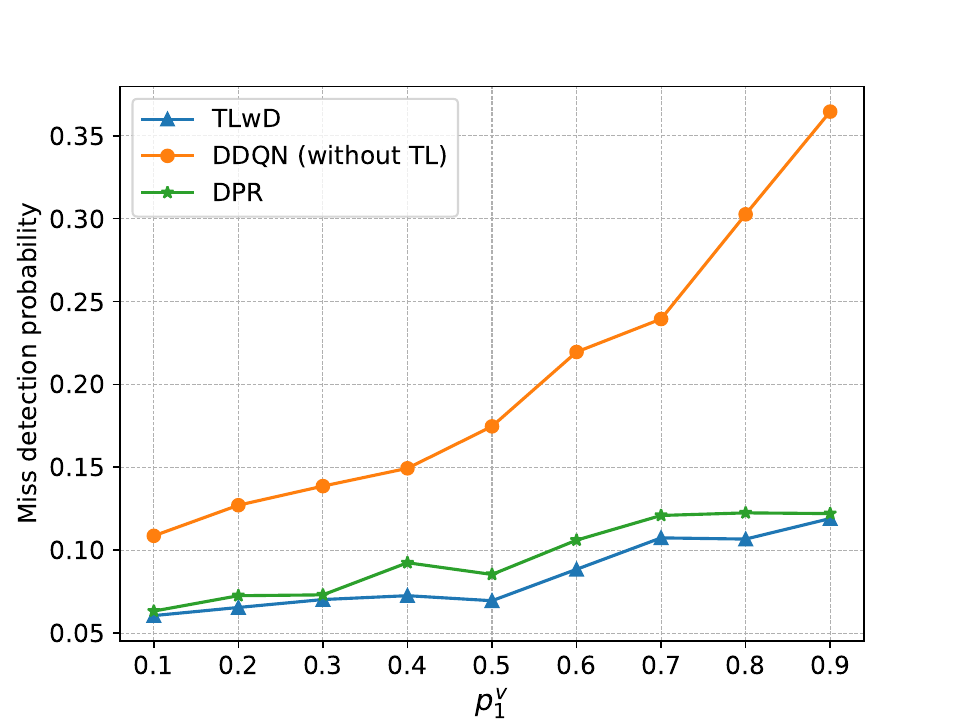}
         \caption{}
     \end{subfigure}
\caption{(a) Average reward, (b) throughput, and (c) miss detection probability vs. probability that unexpected event occurs with high speed of AV ($p_1^v$).}
\label{fig:p1v-varies}
\end{figure*}

\begin{figure*}[t]
\centering
	 \begin{subfigure}[b]{0.35\textwidth}
         \centering
         \includegraphics[width=\textwidth]{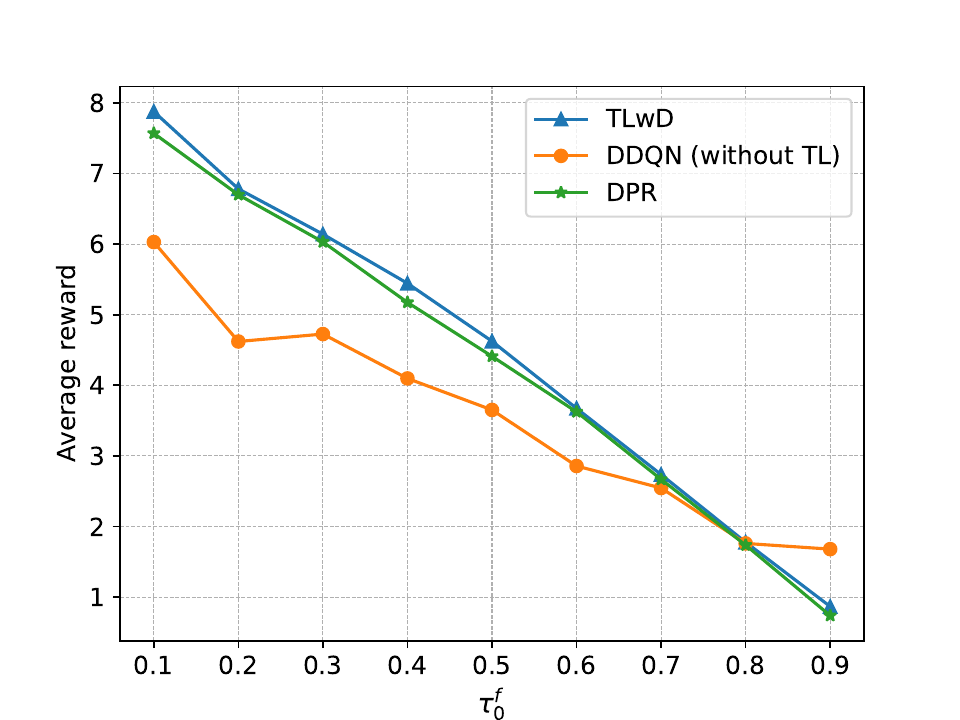}
         \caption{}
     \end{subfigure}
     \hspace{-0.8cm}
     \begin{subfigure}[b]{0.35\textwidth}
         \centering
         \includegraphics[width=\textwidth]{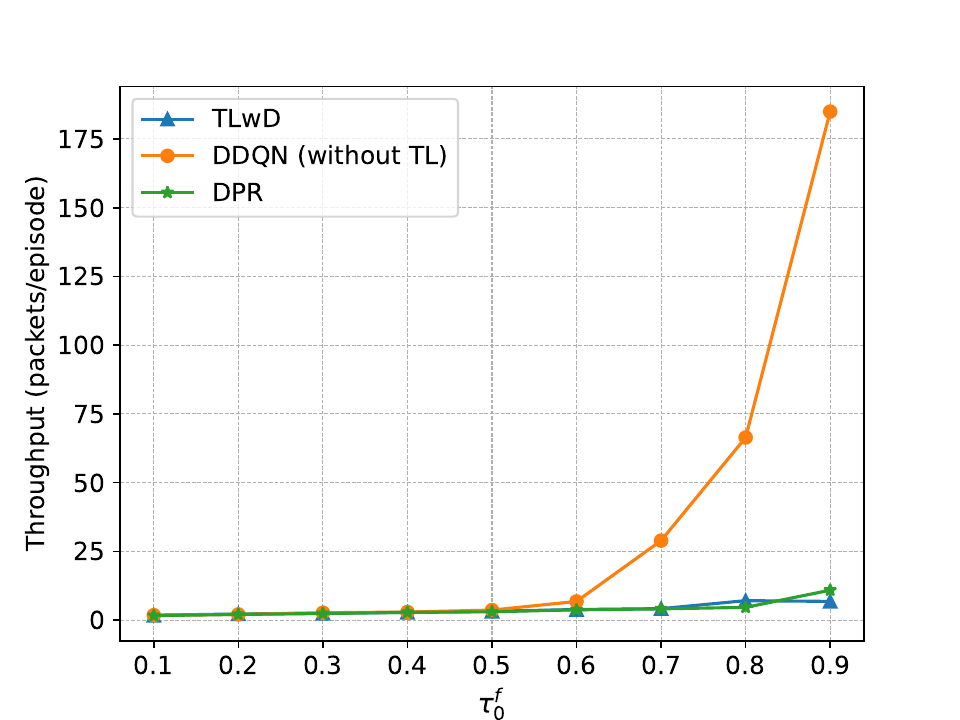}
         \caption{}
     \end{subfigure}
     \hspace{-0.8cm}
     \begin{subfigure}[b]{0.35\textwidth}
         \centering
         \includegraphics[width=\textwidth]{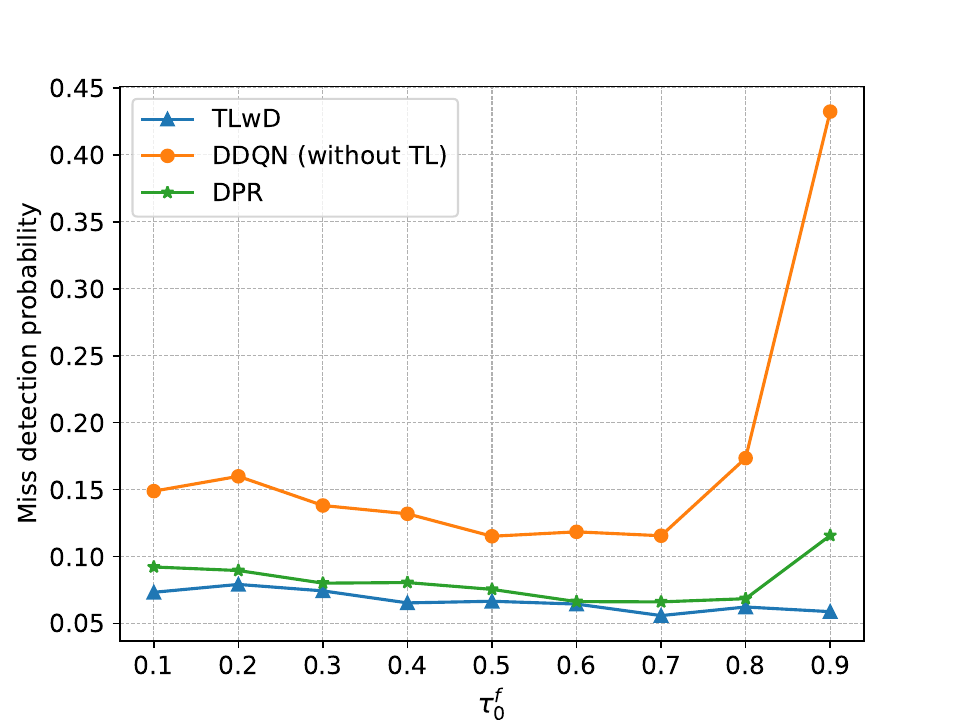}
         \caption{}
     \end{subfigure}
\caption{(a) Average reward, (b) throughput, and (b) miss detection probability vs. probability that the factor $f$ is in favorable condition ($\tau^f_0$).}
\label{fig:tauf_varies}
\end{figure*}

\begin{figure*}[t]
\centering
	 \begin{subfigure}[b]{0.35\textwidth}
         \centering
         \includegraphics[width=\textwidth]{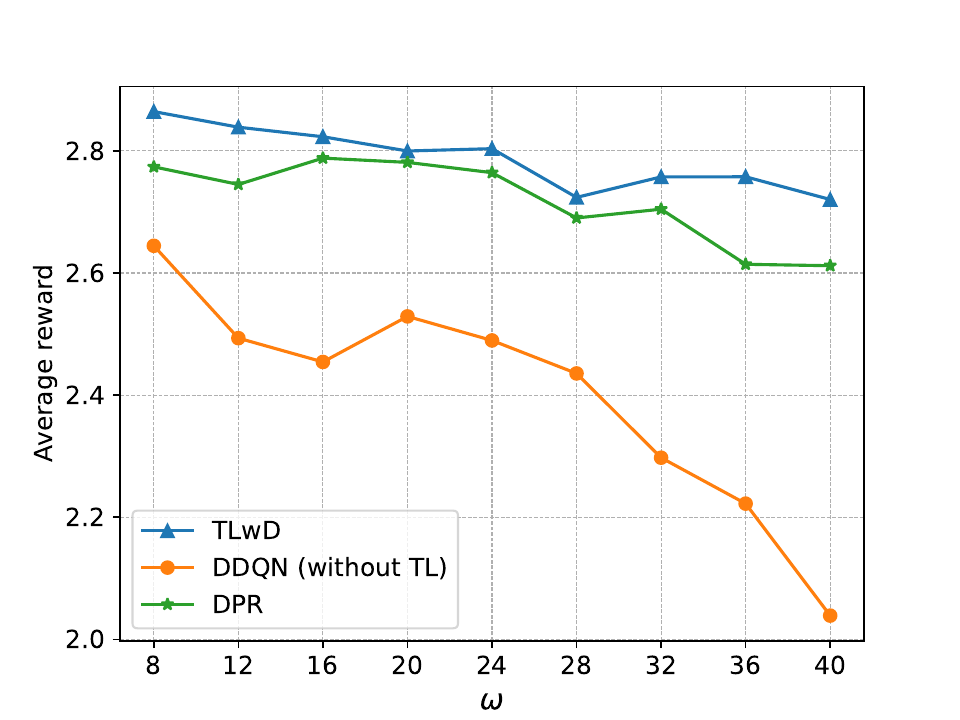}
         \caption{}
     \end{subfigure}
     \hspace{-0.8cm}
     \begin{subfigure}[b]{0.35\textwidth}
         \centering
         \includegraphics[width=\textwidth]{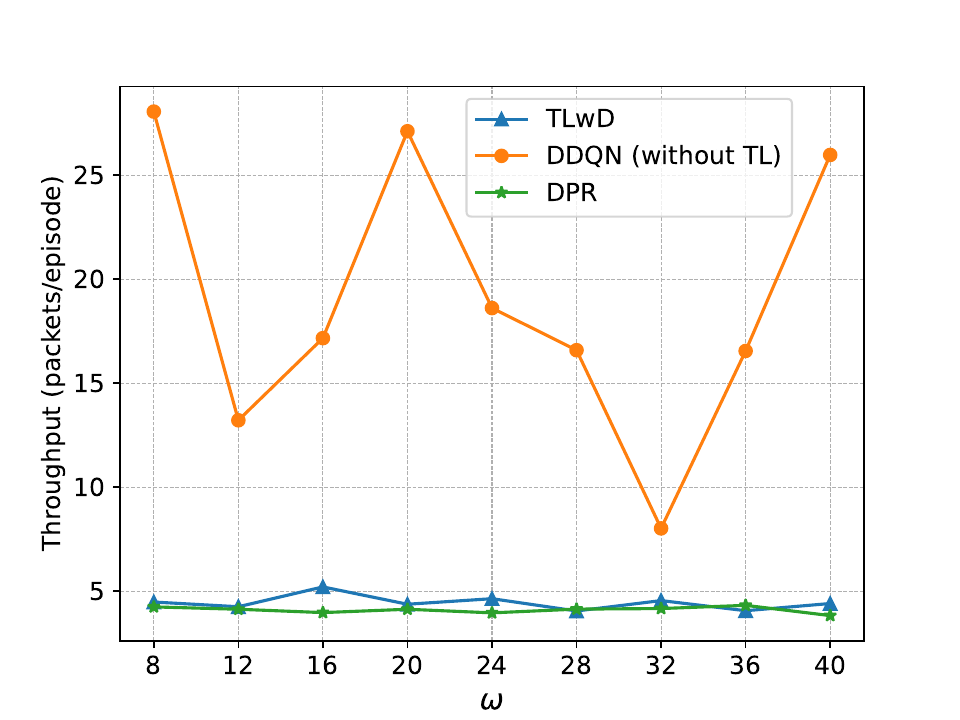}
         \caption{}
     \end{subfigure}
     \hspace{-0.8cm}
     \begin{subfigure}[b]{0.35\textwidth}
         \centering
         \includegraphics[width=\textwidth]{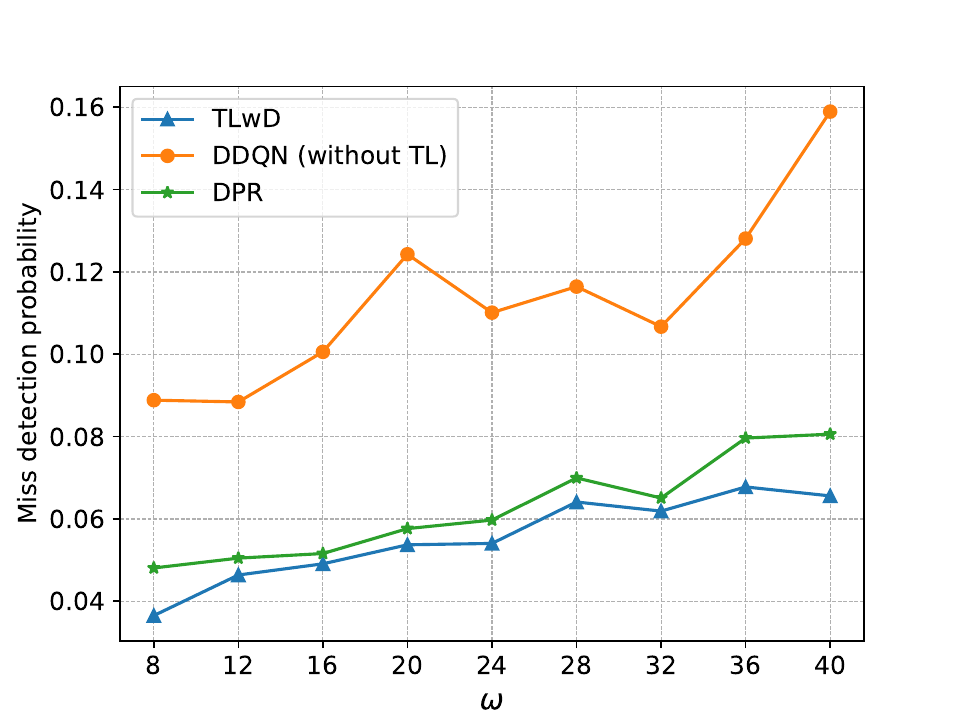}
         \caption{}
     \end{subfigure}
\caption{(a) Average reward, (b) throughput, and (c) miss detection probability vs. average traffic density ($\omega$).}
\label{fig:nt_varies}
\end{figure*}

To further evaluate the robustness of the proposed TLwD algorithm, we vary the probability that an unexpected event occurs when the AV moves at a high speed, i.e., $p_1^v$ in (\ref{eq:unexpected-ev-prob}), and show the results of the three algorithms in Fig.~\ref{fig:p1v-varies}. It is worth noting that each point in Fig.~\ref{fig:p1v-varies},~\ref{fig:tauf_varies}, and~\ref{fig:nt_varies} is obtained by averaging metric values in the first 1000 episodes, including the exploration when the value of $\epsilon$ in $\epsilon$-greedy scheme is decreased from 1.0 to 0.01.
As shown in Fig.~\ref{fig:p1v-varies}(a), when $p_1^v$ increases from 0.1 to 0.9, the TLwD outperforms DPR and DDQN in terms of the average reward. The reason for the upward trend of TLwD and DPR is that when $p_1^v$ increases, the TLwD and DPR algorithms select the radar detection mode more frequently to avoid miss detection penalties and obtain greater rewards. With the values of $p_1^v$ from 0.6 to 0.9, we can observe that the average reward values of DDQN decrease significantly. The reason is that with the high values of $p_1^v$, DDQN takes more episodes to achieve the optimal policy compared to those of the TLwD and DPR, thus resulting in lower average reward values. For example, with $p_1^v = 0.8$, DDQN takes 400 episodes to converge to the optimal policy, while TLwD and DPR take around 150 episodes to achieve similar reward values. Moreover, the average reward values of TLwD are slightly higher than that of DPR because in the early episodes, TLwD can outperform DPR as discussed in Fig.~\ref{fig:reward-results}(b).
The results of throughput and miss detection probability of the three algorithms are shown in Fig.~\ref{fig:p1v-varies}(b) and Fig.~\ref{fig:p1v-varies}(c), respectively. Note that since the AV is able to perform either radar or data communication at each time step, there is a tradeoff between the throughput and miss detection probability. In particular, a higher throughput results in a higher miss detection probability and vice versa. 
In Fig.~\ref{fig:p1v-varies}(b) and Fig.~\ref{fig:p1v-varies}(c), we can observe that TLwD obtains more stable results for miss detection probability and throughput by following the policy that minimizes both miss detection probability and throughput. 
With the DPR, the achieved miss detection probability is higher than that of TLwD. Without using any transfer learning technique, the DDQN algorithm obtains high throughput and high miss detection probability at the same time. In particular, with this transfer learning setting, the miss detection probability obtained by the proposed TLwD algorithm is much lower than that of the conventional DDQN algorithm, reduced by up to $67\%$ as shown in Fig.~\ref{fig:p1v-varies}(c).

Next, we also evaluate the performance of the proposed TLwD algorithm by varying the probability that the factor $f$ is at state 0 (favorable condition), i.e., $\tau^f_0$ in (\ref{eq:unexpected-ev-prob}). In particular, a higher value of $\tau^f_0$ results in a higher probability that an environment state (e.g., communication channel state, and weather state) is at a favorable condition (e.g., good communication channel condition and good weather condition).
Without loss of generality, we increase all the values of ($\tau^r_0, \tau^w_0, \tau^v_0, \tau^m_0, \tau^c_0$) at the same time, where $\tau^r_0$, $\tau^w_0$, $\tau^v_0$, and $\tau^m_0$ are defined in (\ref{eq:unexpected-ev-prob}), and $\tau^c_0$ is the communication channel switching probability. We observe from Fig.~\ref{fig:tauf_varies}(a) that when $\tau^f_0$ increases from 0.1 to 0.9, average reward values of the three algorithms reduce significantly. However, the results of the TLwD algorithm are slightly better than that of DPR and much higher than that of DDQN, especially with the low values of $\tau^f_0$, i.e., $\tau^f_0 < 0.7$. The reason for this downward trend is that the increase of $\tau^f_0$ results in reducing the probability of an unexpected event to occur. With the low values of $\tau^f_0$, meaning that unexpected events occur more frequently, the TLwD and DPR algorithms significantly outperform the DDQN algorithm, thanks to the transfer learning techniques. In contrast, with high values of $\tau^f_0$, i.e., $\tau^f_0 \geq 0.7$, meaning that unexpected events are less likely to occur, the performance gaps among TLwD, DPR, and DDQN are shortened. With $\tau^f_0 = 0.9$, we observe that the average reward of DDQN is better than those of TLwD and DPR, thanks to the policy of DDQN that significantly increases the throughput and thus increases the miss detection probability at the same time. 
The above results also reveal an interesting finding that transfer learning is more efficient when the system's dynamic is considerably high, which is equivalent to the scenarios with small values of $\tau^f_0$. When the system's dynamic is low, e.g., less unexpected events occur, the transfer learning might not be effective as DDQN.
In Fig.~\ref{fig:tauf_varies}(b), with the high values of $\tau^f_0$, i.e., $\tau^f_0 \geq 0.7$, the throughput of DDQN increases significantly while the throughputs of the TLwD and DPR algorithms slightly increase. Consequently, the miss detection probability of DDQN increases when $\tau^f_0 \geq 0.7$ and the miss detection probabilities of TLwD and DPR algorithms still remain at low values.   

In Fig.~\ref{fig:nt_varies}, we vary the average traffic density $\omega$ and evaluate the performance of the proposed TLwD algorithm. Fi.~\ref{fig:nt_varies}(a) shows that the average rewards of the three algorithms decrease when $\omega$ increases from 8 to 40, and the average rewards obtained by TLwD are always higher than those of DPR and DDQN. The reason is that the three algorithms are likely to follow the policies that select the radar detection mode more frequently. As a result, this leads to the decrease of immediate reward obtained from the third and the fourth conditions in (\ref{eq:reward}). Accordingly, the miss detection probability values are increased as shown in Fig.~\ref{fig:nt_varies}(c). In particular, the TLwD algorithm can reduce the miss detection probability by up to $22\%$ and $61\%$ compared to the DPR and DDQN algorithms, respectively.
The results for the throughput in Fig.~\ref{fig:nt_varies}(b) are similar to the  above scenarios (i.e., Fig.~\ref{fig:p1v-varies}(b) and Fig.~\ref{fig:tauf_varies}(b)) in which the TLwD and DPR algorithms achieve lower throughputs than that of DDQN.

\subsubsection{Benefits Of Prioritized-experience Replay Mechanism}
\begin{figure}
\centering
\includegraphics[width=0.8\linewidth]{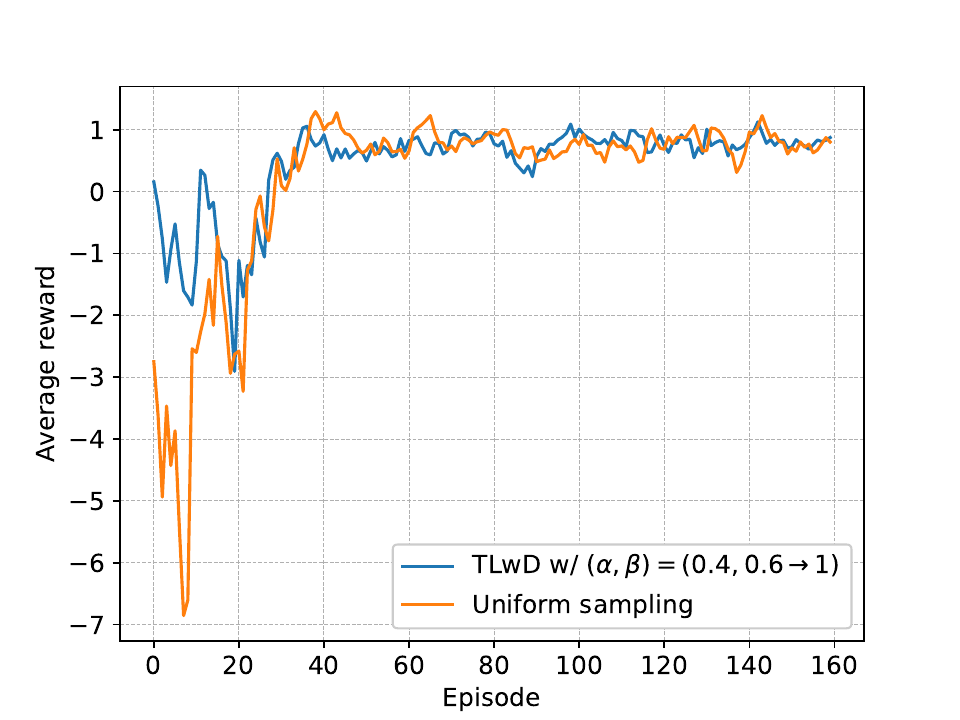}
\caption{Average rewards of TLwD with $(\alpha, \beta) = (0.4, 0.6 \rightarrow 1.0)$ and TLwD with $(\alpha, \beta) = (0, 0)$ (i.e., uniform sampling).}
\label{fig:alpha_varies}
\end{figure}

As discussed in the previous results, the prioritized-experience replay mechanism has a significant impact on the performance of the TLwD algorithm. 
We then evaluate the performance of the TLwD algorithm with and without the prioritized-experience replay mechanism. In particular, the prioritized-experience replay mechanism can be controlled by adjusting two parameters $\alpha$ and $\beta$ in (\ref{eq:prioritized-replay}) and (\ref{eq:weighted-importance-sampling}).
For our prioritized-experience replay mechanism in the TLwD algorithm, we use the common setting in which the value of $\alpha$ is fixed at 0.4 and the value of $\beta$ is increased from 0.6 to 1.0~\cite{schaul2015}, denoted as $(\alpha, \beta) = (0.4, 0.6 \rightarrow 1.0)$. We denote that with the values $(\alpha, \beta) = (0, 0)$, the TLwD algorithm uses uniform sampling mechanism in which the transitions in the replay memory are sampled randomly. Fig.~\ref{fig:alpha_varies} shows that the TLwD algorithm with the prioritized-experience replay mechanism (blue line) acquires faster convergence rate than that of the TLwD algorithm with the uniform sampling mechanism (orange line). This result can confirm the advantage of using prioritized-experience replay technique over traditional uniform sampling technique~\cite{schaul2015, hessel2018}.

\section{Conclusions}
\label{sec:conclusions}

In this paper, we have developed the dynamic framework for joint radar-communications system of autonomous vehicles under the dynamic and uncertainty of surrounding environment. In particular, the dynamic of environment such as communication channel, weather, road and traffic conditions can be captured through the MDP framework and then the optimal policy for the AV can be obtained by using the proposed DDQN algorithm. 
Furthermore, we have developed a transfer learning approach, integrating with the proposed DDQN algorithm, that enables the AV to leverage valuable experiences from similar domains, and thereby significantly improving the AV's system performance when it moves to a new environment. 
Extensive simulations have demonstrated that the transfer learning approach has the convergence rate much faster than those of the conventional deep reinforcement learning approaches in the transfer learning setting.

\begin{IEEEbiography}[{\includegraphics[width=1in,height=1.25in,clip,keepaspectratio]{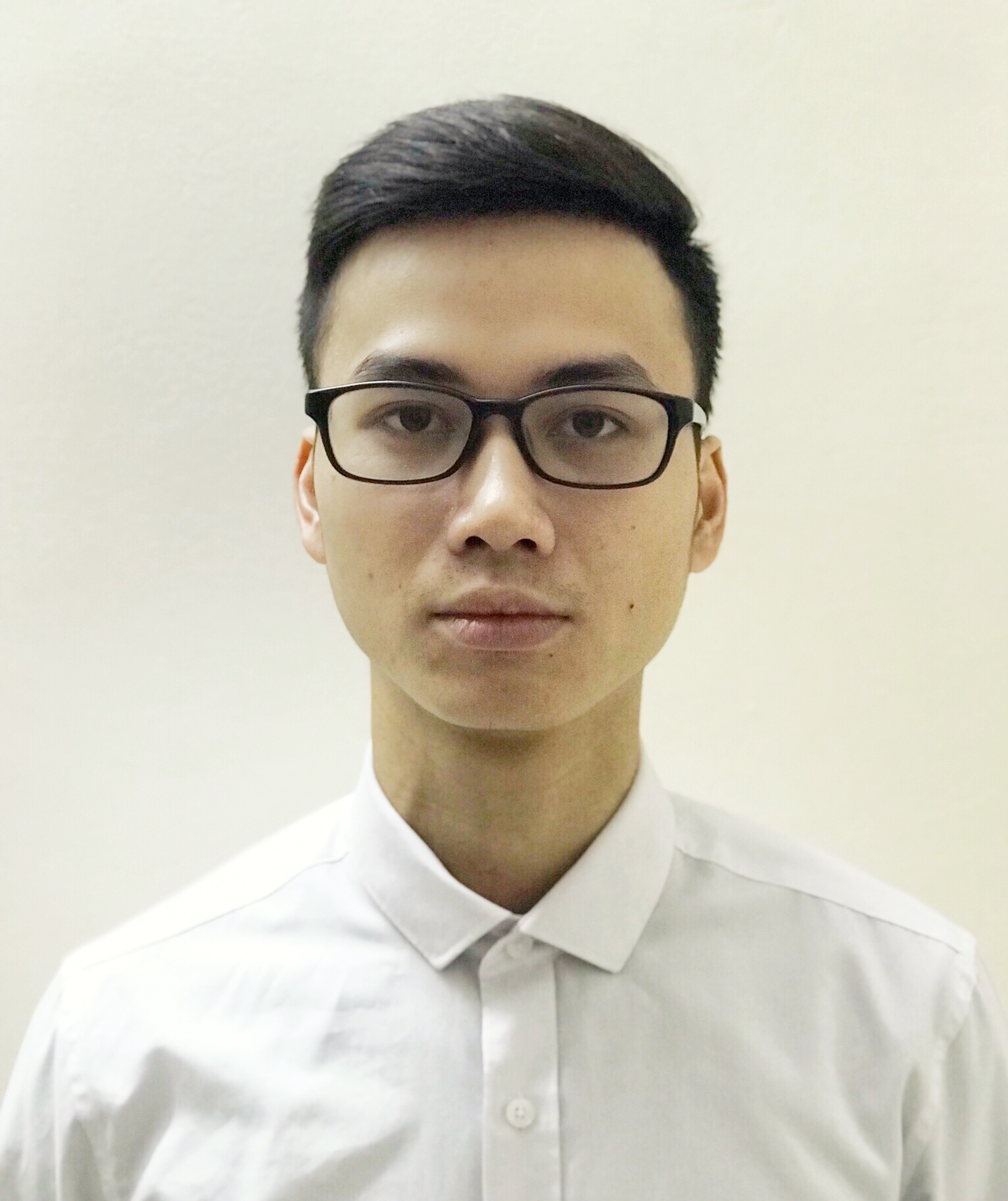}}]%
{Nguyen Quang Hieu} received the B.Eng. degree in Hanoi University of Science Technology, Vietnam in 2018. He is a Ph.D. student in University of Technology Sydney, Australia. He worked as an research assistant in Nanyang Technological University, Singapore from 2019 to 2021. His research interests include wireless communications, network optimization, stochastic optimization, reinforcement learning, and deep learning.  
\end{IEEEbiography}

\vspace{-3.5cm}
\begin{IEEEbiography}[{\includegraphics[width=1in,height=1.25in,clip,keepaspectratio]{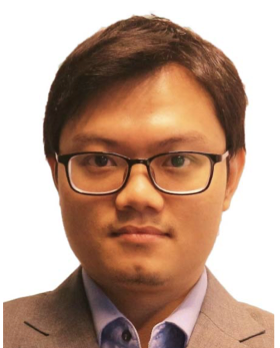}}]%
{Dinh Thai Hoang} (Member, IEEE) received the Ph.D. degree in computer science and engineering from Nanyang Technological University, Singapore, in 2016. He is currently a Faculty Member with the School of Electrical and Data Engineering, University of Technology Sydney, Sydney, NSW, Australia. His research interests include emerging topics in wireless communications and networking, such as machine learning, ambient backscatter communications, IRS, edge intelligence, cybersecurity, the IoT, and 5G/6G networks. He received several awards, including the Australian Research Council. He is also an Editor of IEEE TRANSACTIONS ON WIRELESS COMMUNICATIONS, IEEE TRANSACTIONS ON COGNITIVE COMMUNICATIONS AND NETWORKING, and IEEE WIRELESS COMMUNICATIONS LETTERS, and an Associate Editor of IEEE COMMUNICATIONS SURVEYS AND TUTORIALS.
\end{IEEEbiography}

\vspace{-3.4cm}
\begin{IEEEbiography}[{\includegraphics[width=1in,height=1.25in,clip,keepaspectratio]{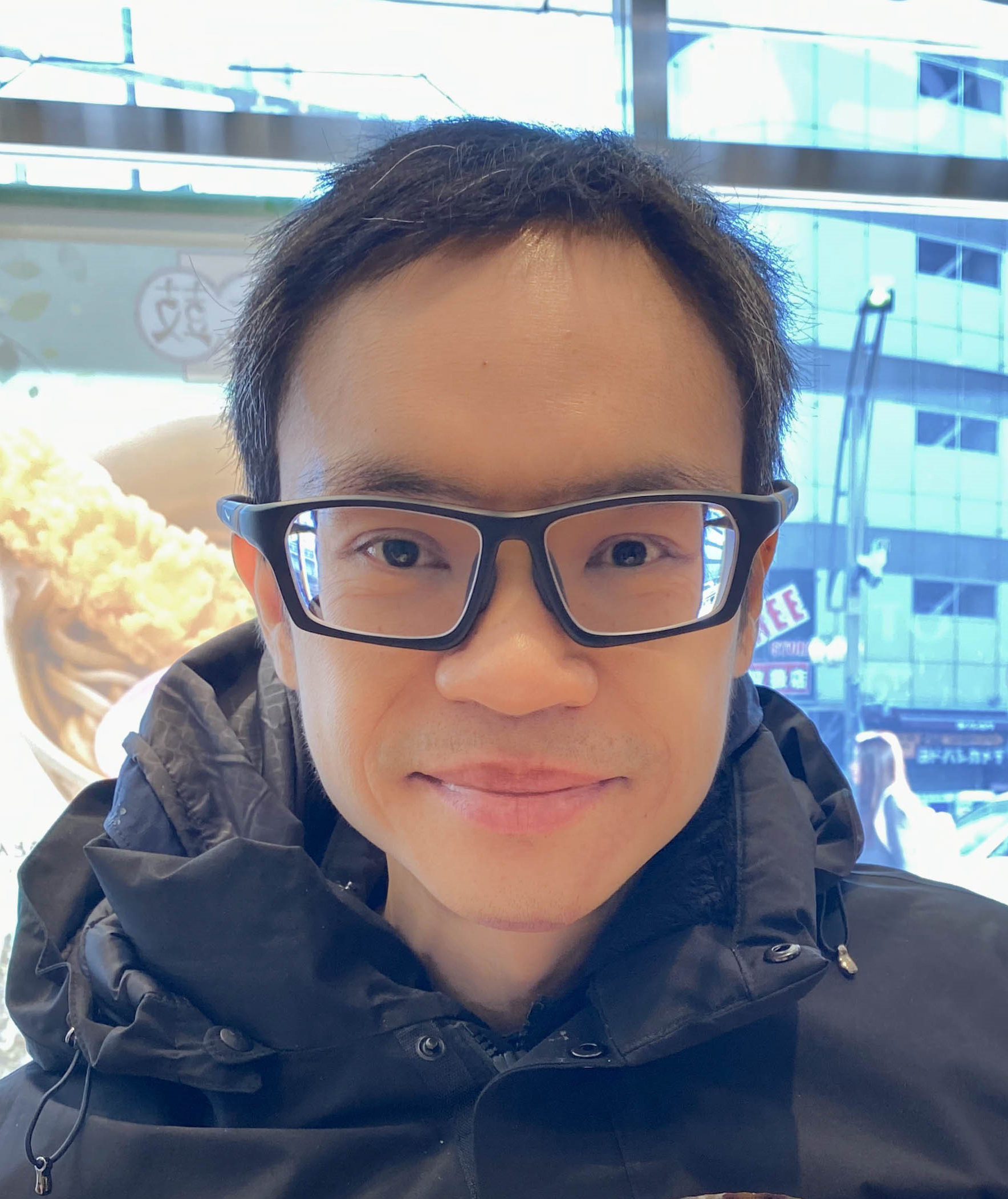}}]%
{Dusit Niyato} (Fellow, IEEE) is a professor in the School of Computer Science and Engineering, at Nanyang Technological University, Singapore. He received B.Eng. from King Mongkuts Institute of Technology Ladkrabang (KMITL), Thailand in 1999 and Ph.D. in Electrical and Computer Engineering from the University of Manitoba, Canada in 2008. His research interests are in the areas of Internet of Things (IoT), machine learning, and incentive mechanism design.
\end{IEEEbiography}

\vspace{-3.4cm}
\begin{IEEEbiography}[{\includegraphics[width=1in,height=1.25in,clip,keepaspectratio]{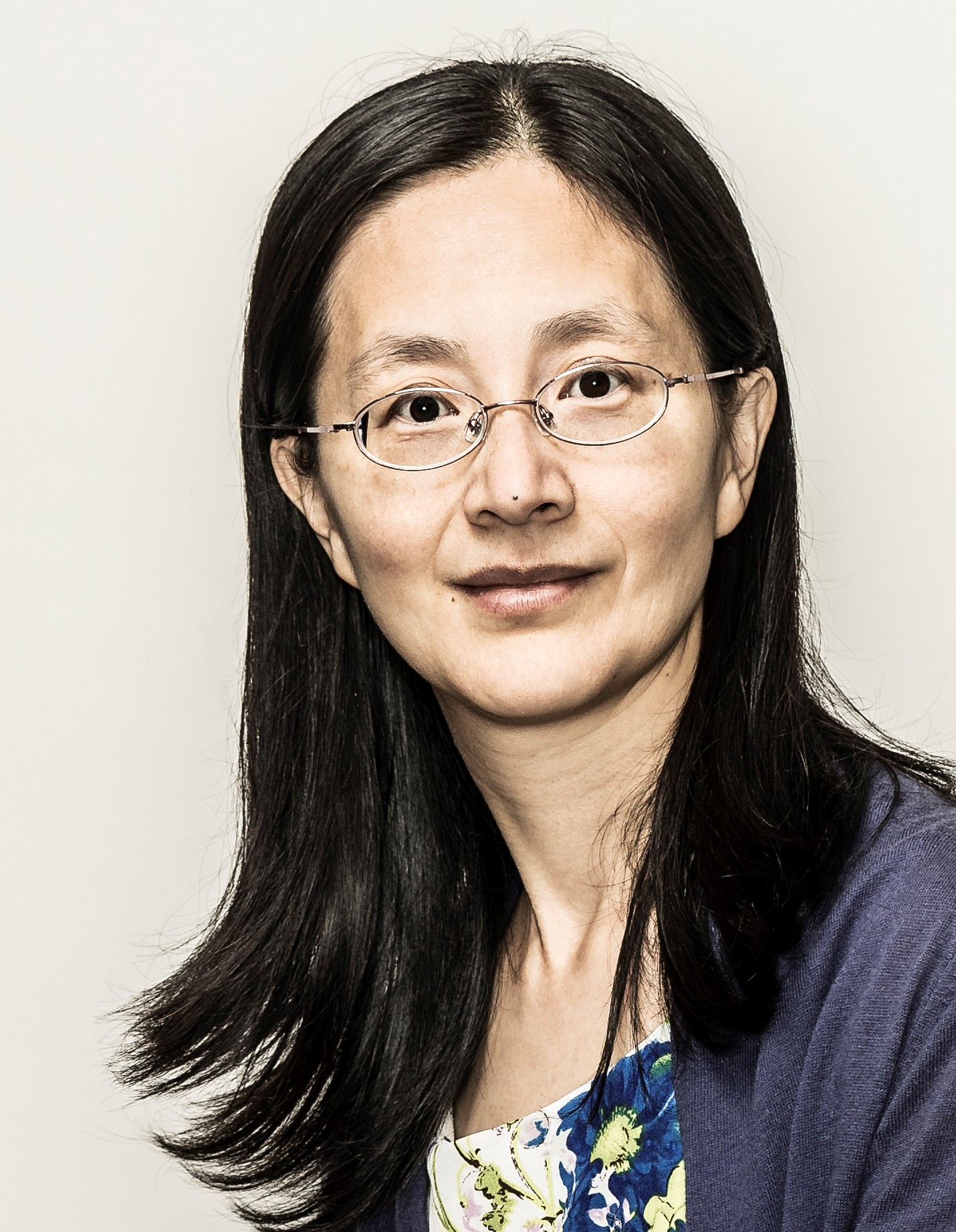}}]%
{Ping Wang}(Fellow, IEEE) is an Associate Professor at the Department of Electrical Engineering and Computer Science, York University, and a Tier 2 York Research Chair. Prior to that, she worked with Nanyang Technological University, Singapore, from 2008 to 2018. Her research interests are mainly in the area of wireless communication networks, cloud computing and Internet of Things with the recent focus on integrating Artificial Intelligence (AI) techniques into communications networks. She has published more than 250 papers/conference proceedings papers. Her scholarly works have been widely disseminated through top-ranked IEEE journals/conferences and received the Best Paper Awards from IEEE Wireless Communications and Networking Conference (WCNC) in 2022, 2020 and 2012, from IEEE Communication Society: Green Communications \& Computing Technical Committee in 2018, and from IEEE International Conference on Communications (ICC) in 2007. Her work received 21,000+ citations with H-index 70 (Google Scholar). She is an IEEE Fellow and a Distinguished Lecturer of the IEEE Vehicular Technology Society. 
\end{IEEEbiography}
\enlargethispage{-0.3cm}

\newpage

\begin{IEEEbiography}[{\includegraphics[width=1in,height=1.25in,clip,keepaspectratio]{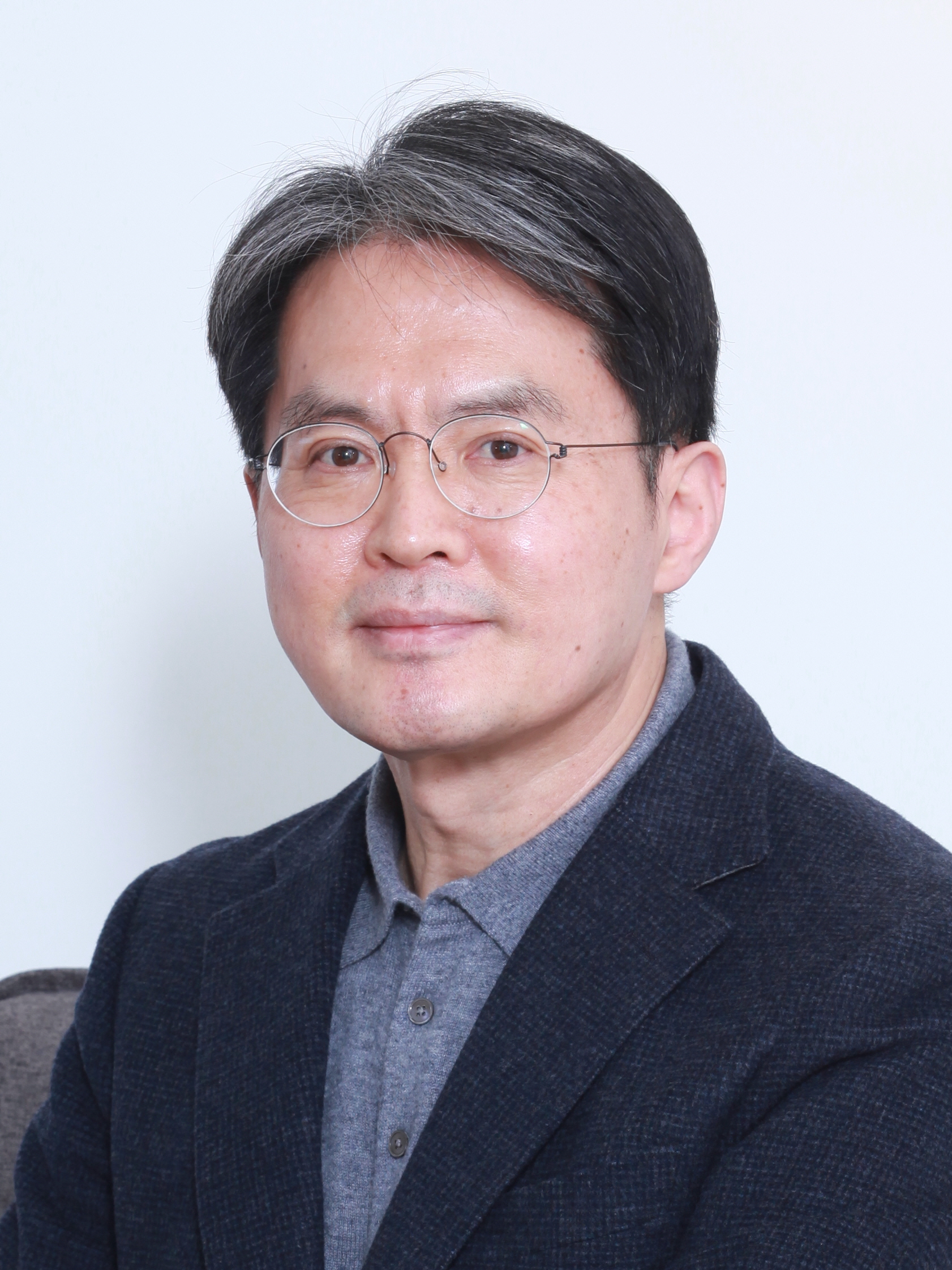}}]%
{Dong In Kim} (Fellow, IEEE) received the Ph.D. degree in electrical engineering from the University of Southern California, Los Angeles, CA, USA, in 1990. He was a Tenured Professor with the School of Engineering Science, Simon Fraser University, Burnaby, BC, Canada. Since 2007, he has been an SKKU-Fellowship Professor with the College of Information and Communication Engineering, Sungkyunkwan University (SKKU), Suwon, South Korea. He is a Fellow of the Korean Academy of Science and Technology and a Member of the National Academy of Engineering of Korea. He has been a first recipient of the NRF of Korea Engineering Research Center in Wireless Communications for RF Energy Harvesting since 2014. He has been listed as a 2020 Highly Cited Researcher by Clarivate Analytics. From 2001 to 2020, he served as an editor and an editor at large of Wireless Communications I for the IEEE Transactions on Communications. From 2002 to 2011, he also served as an editor and a Founding Area Editor of Cross-Layer Design and Optimization for the IEEE Transactions on Wireless Communications. From 2008 to 2011, he served as the Co-Editor-in-Chief for the IEEE/KICS Journal of Communications and Networks. He served as the Founding Editor-in-Chief for the IEEE Wireless Communications Letters, from 2012 to 2015. He was selected the 2019 recipient of the IEEE Communications Society Joseph LoCicero Award for Exemplary Service to Publications. He is the General Chair for IEEE ICC 2022 in Seoul.
\end{IEEEbiography}

\vspace{15cm}
\begin{IEEEbiography}[{\includegraphics[width=1in,height=1.25in,clip,keepaspectratio]{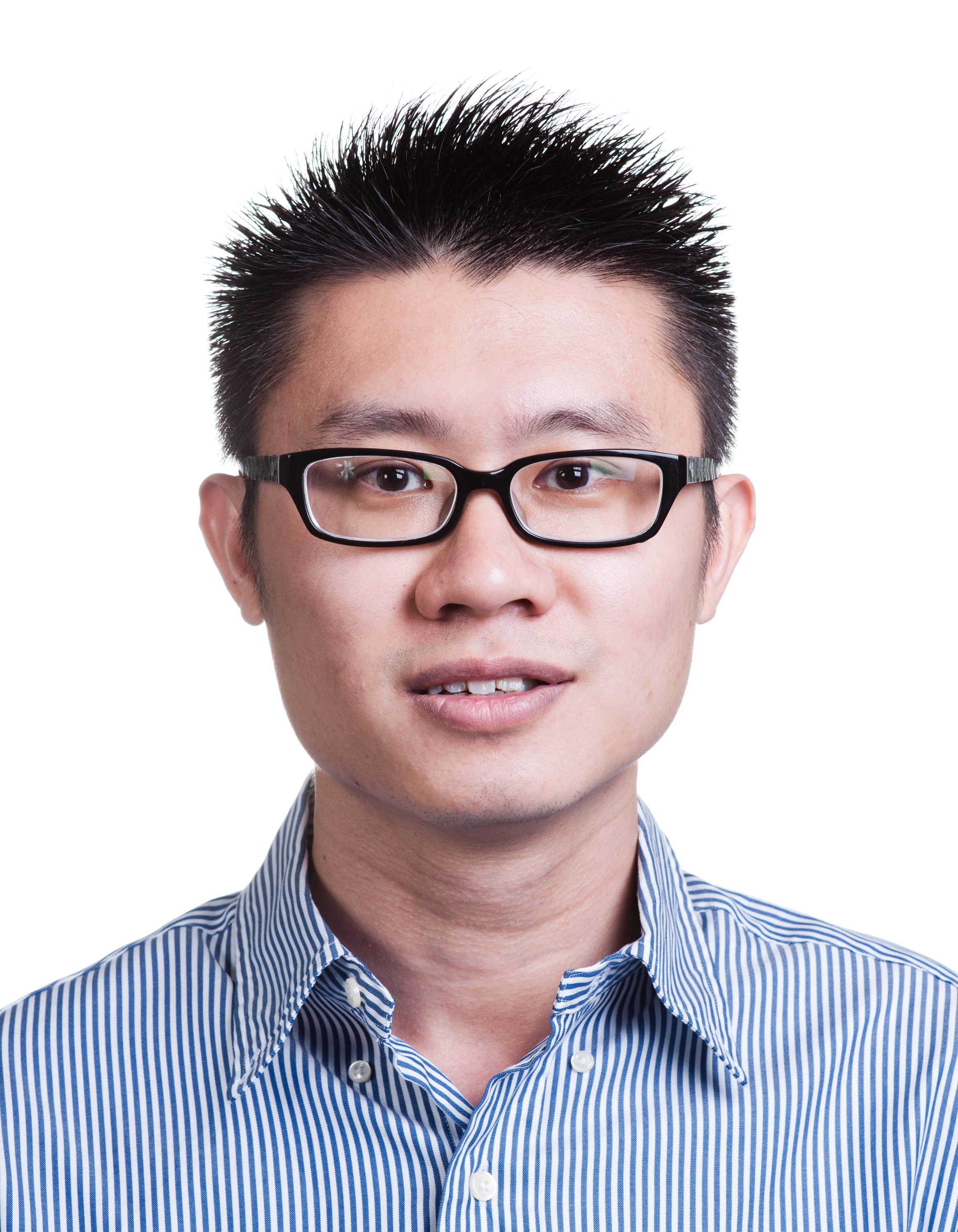}}]%
{Chau Yuen } (Fellow, IEEE) received the B.Eng. and Ph.D. degrees from Nanyang Technological University (NTU), Singapore, in 2000 and 2004, respectively. He was a Post-Doctoral Fellow with Lucent Technologies Bell Labs, Murray Hill, in 2005. From 2006 to 2010, he was with the Institute for Infocomm Research (I2R), Singapore. Since 2010, he has been with the Singapore University of Technology and Design. Dr. Yuen was a recipient of the Lee Kuan Yew Gold Medal, the Institution of Electrical Engineers Book Prize, the Institute of Engineering of Singapore Gold Medal, the Merck Sharp and Dohme Gold Medal, and twice a recipient of the Hewlett Packard Prize. He received the IEEE Asia Pacific Outstanding Young Researcher Award in 2012 and IEEE VTS Singapore Chapter Outstanding Service Award on 2019. Currently, he serves as an Editor for the IEEE TRANSACTIONS ON VEHICULAR TECHNOLOGY, IEEE System Journal, and IEEE Transactions on Network Science and Engineering . He served as the guest editor for several special issues, including IEEE JOURNAL ON SELECTED AREAS IN COMMUNICATIONS, IEEE WIRELESS COMMUNICATIONS MAGAZINE, IEEE TRANSACTIONS ON COGNITIVE COMMUNICATIONS AND NETWORKING. He is a Distinguished Lecturer of IEEE Vehicular Technology Society.
\end{IEEEbiography}
\enlargethispage{-4.5cm}

\end{document}